%% file: main.tex
\documentclass[runningheads]{llncs}

\usepackage{eccv}

\usepackage{eccvabbrv}

\usepackage{graphicx}
\usepackage{booktabs}

\usepackage[accsupp]{axessibility}  % Improves PDF readability for those with disabilities.

\usepackage{hyperref}

\usepackage{orcidlink}

\usepackage{caption}
\usepackage{multirow}

\usepackage{pifont}% http://ctan.org/pkg/pifont
\newcommand{\cmark}{\ding{51}}%
\newcommand{\xmark}{\ding{55}}%

\newcommand{\OURS}{EmoteGPT\xspace}
\newcommand{\DATASET}{Txt2Emote\xspace}

\begin{document}

% \title{Author Guidelines for ECCV Submission} 
\title{\OURS: 3D Human Facial Expressions from Natural Language Descriptions}

\titlerunning{\OURS}

\author{%
Haoran Wang\inst{1} \and
% \orcidlink{0000-0001-6177-9706} \and
 Mohit Mendiratta\inst{1} \and \\
Christian Theobalt \inst{1} \and
Adam Kortylewski\inst{2}
% \orcidlink{0000-0002-9146-4403}
}

\authorrunning{H.~Wang et al.}

\institute{Max Planck Institute for Informatics, Saarland Informatics Campus \and
CISPA Helmholtz Center for Information Security}

\maketitle

\begin{center}
\vspace{-8pt}
{\small \textbf{Project Page:} \url{https://genintel.github.io/EmoteGPT}}
\vspace{10pt}
\end{center}

\input{sec/0_abstract}

\begin{figure}[ht]
  \centering
   \includegraphics[width=\linewidth]{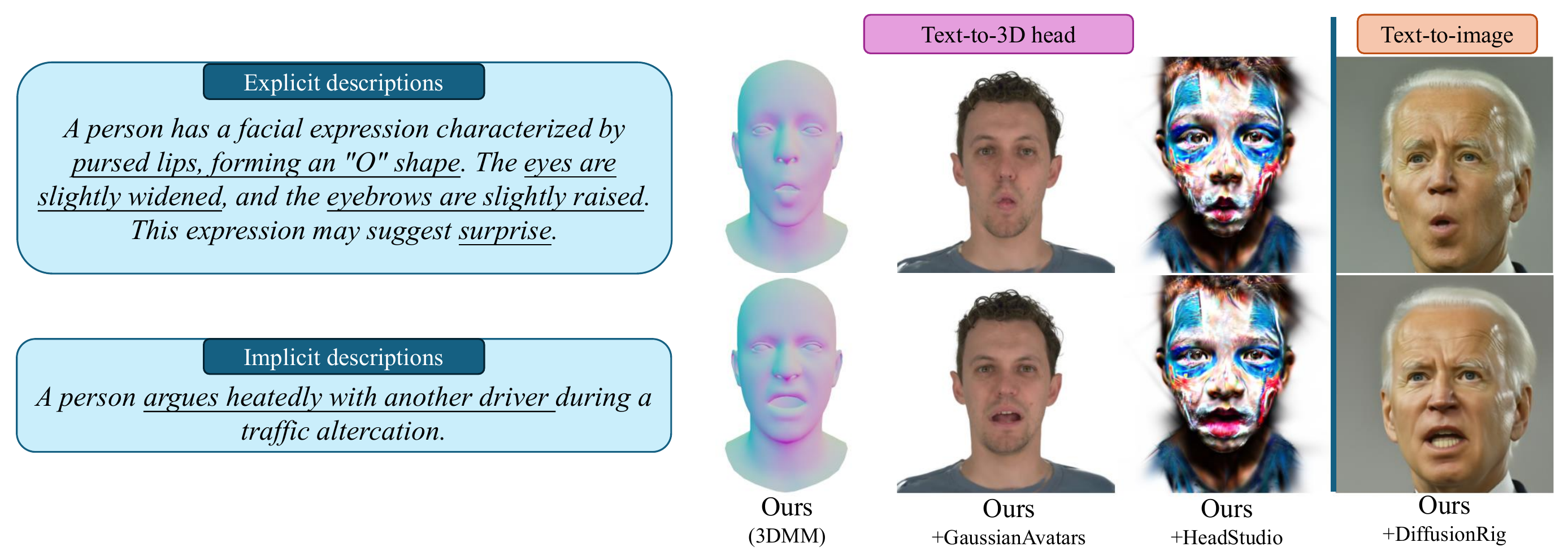}

   \caption{
   \OURS generates 3D facial expressions from text, supporting two types of descriptions: 
    (1) \textit{Explicit} ones, which detail physical facial features and overall emotional impressions; (2) \textit{Implicit} ones, which reference events or situations that evoke facial expressions.
The expressions are represented as 3DMM parameters and can be seamlessly integrated into existing frameworks~\cite{qian2024gaussianavatars, zhou2024headstudio,ding2023diffusionrig} to enable expressive 3D avatars and personalized 2D face synthesis.
   }
   \label{fig:teaser_figure}
\end{figure}

\input{sec/1_intro}

\input{sec/2_related_works}

\input{sec/3_dataset}

\input{sec/3_method}

\input{sec/4_experiments}

\input{sec/5_limitations}
\input{sec/5_conclusion}

\section*{Acknowledge}
Adam Kortylewski acknowledges support via his Emmy Noether Research Group funded by the German Research Foundation (DFG) under Grant No. 468670075.

% \clearpage\mbox{}Page \thepage\ of the manuscript.
% \clearpage\mbox{}Page \thepage\ of the manuscript.
% \clearpage\mbox{}Page \thepage\ of the manuscript.
% \clearpage\mbox{}Page \thepage\ of the manuscript.
% \clearpage\mbox{}Page \thepage\ of the manuscript. This is the last page.
% Now we have reached the maximum length of an ECCV \ECCVyear{} submission (excluding references and acknowledgements).
% References should start immediately after the main text, but can continue past p.\ 14 if needed. 
\clearpage  % TODO FINAL: This \clearpage needs to be removed from both review and camera-ready versions.

% \section*{Acknowledgements}
% Please insert your acknowledgments here.

\bibliographystyle{splncs04}
\bibliography{main}

\input{sec/X_suppl}
\end{document}

%% file: sec/0_abstract.tex
\begin{abstract}
Precise control of 3D facial expressions from text is crucial for virtual avatars, animation, and human–computer interaction, yet existing text-to-3D methods jointly generate identity, expression, and texture, making fine-grained expression control difficult. We instead formulate text-driven expression synthesis as a regression problem in the disentangled parameter space of a 3D Morphable Model (3DMM).
This setting, however, requires paired data linking detailed language to precise expression parameters, which are missing from existing resources.
To fill this gap, we introduce \DATASET, a benchmark of diverse 3D facial expressions with fine-grained textual annotations obtained from GPT-4o and a high-fidelity face tracker, providing both explicit descriptions detailing facial features and implicit descriptions referencing the situational context behind the expression.
Leveraging this dataset, we present \OURS, a text-to-3D expression framework based on a Multimodal Large Language Model (MLLM) with a dedicated \texttt{<Expr>} token to semantically ground expression representations, which are then decoded into 3DMM parameters. 
We further improve \OURS by augmenting training with large-scale image-to-3DMM data, enabling it to surpass state-of-the-art text-to-3D face synthesis methods on emotion recognition metrics and in perceived expressiveness.
Integrated into avatar pipelines, our method enables photorealistic and stylized 3D avatars, as well as expressive 3D-consistent 2D face synthesis from textual input.

\keywords{Visual Language Model \and Face Synthesis}
\end{abstract}

%% file: sec/1_intro.tex
\section{Introduction}
\label{sec:intro}

Controllable 3D facial expression synthesis has broad applications in virtual avatars, animation, and human–computer interaction. Traditional control methods such as blendshape editing are labor-intensive and unintuitive for non-experts. 
In contrast, natural language offers a flexible and universal interface for expression control. Users can describe expressions in diverse ways, using \textbf{explicit visual cues} (e.g., “He is smiling”) or \textbf{implicit contextual statements} (e.g., “He looks like he just won the lottery”). 
Explicit descriptions specify observable facial features, while implicit ones encode the situational or emotional context that evokes those features. Handling both forms is crucial for natural interaction, as people rarely describe emotions by enumerating muscle movements.

Recent works~\cite{10.1145/3618368, magicmirrorfasthighqualityavatar2024} have explored generating expressive 3D heads from text, predominantly using diffusion-based text-to-3D pipelines. However, these methods remain limited for 3D expression control: they typically entangle identity, pose, and expression. They also struggle to capture nuanced and diverse language descriptions, and require expensive iterative sampling that limits efficiency. As a result, existing text-to-3D head generation methods fall short in scenarios demanding accurate, independently controllable, and efficient expression synthesis.

We instead address these challenges by formulating the problem as a \textit{language-to-expression regression} task, where expressions are represented explicitly in the compact and disentangled parameter space of a 3D Morphable Model (3DMM) \cite{blanz1999morphable,egger20203d}. 
This 3DMM-based formulation provides a structured and interpretable representation that decouples expression from identity, enhances controllability and expressive fidelity, and enables efficient inference.

A key obstacle is the lack of data linking natural language to precise 3D expression parameters. To address this, we construct \DATASET, a dataset of 30k 3DMM expressions paired with diverse textual descriptions. Each expression is annotated with two complementary text types: (1) \textbf{explicit} annotations that detail observable facial features (e.g., mouth shape, eye tension), and (2) \textbf{implicit} annotations reflecting emotional or situational context. Figure~\ref{fig:teaser_figure} illustrates representative samples. 
We also reserve 2.5k expressions as a held-out benchmark for evaluating models on both explicit and implicit language.

Building on \DATASET, we propose \OURS, a framework for generating 3D facial expressions from natural language. \OURS uses a Multimodal Large Language Model (MLLM) to directly regress 3DMM expression parameters. We introduce a dedicated \texttt{<Expr>} token whose representation can be decoded by a lightweight expression head into 3DMM parameters, yielding deterministic, real-time predictions without diffusion-style iterative sampling. Operating in the 3DMM expression space allows \OURS to focus exclusively on expression, enabling precise and controllable manipulation while remaining compatible with existing avatar systems.

\OURS further supports multimodal supervision, leveraging large-scale image-to-expression data and instruction-following text data to strengthen the alignment between language and facial expression. 
The resulting model integrates seamlessly with existing 3DMM-based avatar pipelines~\cite{qian2024gaussianavatars, wei2023dediffusion, zhou2024headstudio}, enabling a wide range of applications including photorealistic avatar synthesis, stylized avatar generation, and personalized facial modeling (Figure~\ref{fig:teaser_figure}).
Extensive experiments demonstrate that \OURS significantly outperforms prior CLIP- and diffusion-based baselines in both emotion recognition accuracy and expressive fidelity.

Our contributions are summarized as follows:
\begin{itemize}
    \item We propose \OURS, a new method that leverages a Multimodal Large Language Model (MLLM) for generating 3D facial expressions from natural language, enabling precise and expressive facial expression synthesis.

    \item We introduce \DATASET, a new dataset of 30k 3D facial expressions paired with fine-grained textual annotations, including \textit{explicit} visual cues and \textit{implicit} contextual text. 

    \item We demonstrate that \OURS benefits from multimodal face data, which strengthens the link between language and facial expression, and that it substantially outperforms state-of-the-art CLIP-based diffusion models and MLLM baselines on both emotion recognition accuracy and expressiveness.
\end{itemize}

%% file: sec/2_related_works.tex
\section{Related work}
\label{sec:related_works}

\textbf{Face datasets with text annotations.}
Several multimodal face datasets with text annotations exist, but most focus on describing static visual attributes or general appearance, offering only limited coverage of facial expressions. CelebA-Dialog~\cite{CelebA-Dialog} provides captions of five attributes for face images in the CelebA dataset where only the smile attribute is related to human facial expressions.
MMCelebA-HQ~\cite{xia2021tedigan} uses an automatic template-based approach to generate captions from attribute annotations in CelebA-HQ, limiting the diversity and expressiveness of the text.
CelebAText~\cite{celebatext2021} provides human-labeled captions for 15k images in CelebA-HQ dataset, yet the annotations primarily emphasize high-level visual attributes or overall appearance rather than nuanced expression details.
While some datasets include expression-related text annotations, such descriptions tend to be coarse, capturing only basic states, e.g. whether the subject is smiling or has an open mouth, without detailing the underlying facial muscle movements or compound emotions. As a result, current multimodal face datasets with textual descriptions remain insufficient for studying or modeling fine-grained facial expressions.

Such limited annotations hinder the ability of models to learn fine-grained mappings between language and facial expression. 
Furthermore, existing datasets often lack contextual or situational cues that might evoke expressions. 
This absence of contextual grounding makes it challenging for generative models to synthesize realistic and expressive faces from text, particularly when the input does not explicitly describe facial expressions.

\textbf{Text-to-3D face generation.}
Early works on text-to-3D face synthesis, such as Describe3D~\cite{describe3d2023}, CLIP-Face~\cite{aneja2023clipface}, and CLIP-Head~\cite{sig2023cliphead}, builds on CLIP to achieve text-guided 3DMM face synthesis by aligning the visual features of rendered faces with the semantic meaning of text prompts. Inspired by DreamFusion~\cite{poole2023dreamfusion}, which demonstrated the potential of Score Distillation Sampling (SDS) for 3D object generation, recent methods~\cite{zhou2024headstudio, Portrait3D_sig24, huang2023humannorm} have started leveraging large pre-trained text-to-image diffusion models for 3D face synthesis from text. However, SDS-based 3D generation often suffers from saturation and noisy artifacts, prompting efforts to address these limitations. For example, HumanNorm~\cite{huang2023humannorm} introduces a multistep SDS pipeline for progressive 3D human mesh generation, while Portrait3D~\cite{Portrait3D_sig24} incorporates a 3D-aware GAN to reduce errors, and HeadStudio~\cite{zhou2024headstudio} uses the FLAME model to improve geometric initialization. Despite these improvements, these methods methods primarily focus on identity and shape while overlooking expressive facial details. 
Some audio-based head generation works~\cite{tan2024style2talker, EMOTE, expclip2024, ma2024talkcliptalkingheadgeneration} incorporate text as additional guidance for expressive synthesis with 3DMMs, but they typically depend on emotion labels or handle only limited, simple descriptions. In contrast, \OURS explicitly bridges diverse and complex text descriptions to expressive 3D face parameters.

\textbf{Multimodal Large Language Models.}
Multi-Modal Large Language Models extend the capabilities of language models beyond text to include a broader spectrum of modalities, such as images, videos, and audio. Recent research in this area enhances their ability to process and integrate multiple modalities, making these models more versatile. In the realm of image-text understanding, approaches like LLaVA~\cite{liu2024visual} and MiniGPT-4~\cite{zhu2023minigpt} incorporate vision encoders to interpret images and align their features with language embeddings using projection layers. Furthermore, models such as PandaGPT~\cite{su2023pandagpt}, ImageBind~\cite{girdhar2023imagebind}, and NeXT-GPT~\cite{wu2023next} show versatility in handling diverse modalities, aligning embedded text, images, audio and video with language as input and output. Approaches such as LISA~\cite{lai2023lisa} connect LLaVA with a decoder to generate text and segmentation masks, while ChatPose~\cite{feng2024chatpose} specializes in human pose information. 
However, there has been no study on integrating MLLMs with 3D human facial expressions, and our work fills this gap.

\textbf{Image-based expressive 3D face reconstruction.}
Expressive 3D face reconstruction has been widely explored in the context of image-based approaches. Existing works~\cite{danvevcek2022emoca, retsinas20243dfacialexpressionsanalysisbyneuralsynthesis} mainly take a single image as input and output 3D facial expressions in the FLAME head model space~\cite{FLAME:SiggraphAsia2017}. EMOCA~\cite{danvevcek2022emoca} incorporates emotion consistency and lip articulation constraints to refine expression accuracy, while SMIRK~\cite{retsinas20243dfacialexpressionsanalysisbyneuralsynthesis} extends MICA~\cite{zielonka2022metricalreconstructionhumanfaces} by leveraging an image-to-image translation network for improved expression synthesis. 
These image-based 3DMM estimation works have been widely integrated in 2D, 3D, or 4D avatar synthesis pipelines~\cite{ding2023diffusionrig, prinzler2024joker, taubner2025cap4d, taubner2025mvp4d}.
While these methods focus on reconstructing expressive 3D faces from images, generating expressive 3D faces directly from text descriptions remains less explored.

%% file: sec/3_dataset.tex
\section{Text-to-3D Expression Dataset}
\label{sec:text2expr_bench}

Aligned text descriptions for diverse facial expressions are necessary to establish a connection between language and 3D facial expressions. However, few public datasets contain text paired with 3D faces. 
As mentioned in Section~\ref{sec:related_works}, existing face datasets with text annotations~\cite{celebatext2021, xia2021tedigan, CelebA-Dialog, ffhqtext} provide limited captions for facial expressions, which hinders research on text-to-3D facial expression generation.
To fill this gap, we introduce \DATASET, a new dataset of expressive faces with individual descriptive captions and identity-isolated 3D meshes.

\input{tables/dataset_comp}
\subsection{Dataset Construction}

We construct our dataset from the AffectNet dataset~\cite{affectnet2019} which contains diverse expressive face images spanning eight emotions.
Importantly, we follow the intuition that humans describe facial expressions in two main ways: (1) \textit{direct} descriptions of facial attributes, such as movements of the eyes, nose, or mouth, and the overall emotion (e.g. ``The face has a smile. This face looks happy''), and (2) \textit{indirect} descriptions via activities or scenarios that could evoke the expression (e.g ``The face looks as if he/she just failed an exam''). 
In this work, we refer to the first type as \textit{explicit} descriptions and the second as \textit{implicit} descriptions. 
For each expressive facial instance in our dataset, we provide a 3D face mesh along with both forms of text:

\textbf{\textit{Explicit descriptions}}: 
    These provide comprehensive details of facial features along with a high-level summary of the associated emotion. Unlike existing datasets, which offer only coarse or partial annotations, our descriptions capture the complexity of expressions by specifying physical attributes in different facial regions and emotional nuances.
    
\textbf{\textit{Implicit descriptions}}: In contrast to existing face datasets, which focus solely on facial features or emotion labels, implicit descriptions depict everyday scenarios or actions that naturally evoke the corresponding facial expression. This offers a more natural, human-centric way of describing expressions and aligns with how people intuitively understand and communicate emotions.

To build the dataset, we first subsample AffectNet according to its emotion annotations to obtain a balanced set with a similar number of images per emotion category. 
For each selected image, we reconstruct a 3D facial mesh using a robust expressive face estimator~\cite{danvevcek2022emoca} and generate both explicit and implicit descriptions with GPT-4o~\cite{gpt4}. 
We then filter the reconstructed meshes and generated text annotations to reduce noise. 
The resulting dataset contains 30k expressive facial images, each paired with a 3D mesh and two text annotations. 
Among them, 27k samples are drawn from the AffectNet training set and 3k from the AffectNet test set. 
Following the AffectNet protocol, we manually inspect the 3k test samples and retain 2.5k high-quality samples as the evaluation benchmark. 
Table~\ref{table:dataset_comp} compares our dataset with existing face datasets containing text annotations. 
Further details on data generation and prompting are provided in Supplementary Sec.~\ref{sec:data_generation}.

%% file: tables/dataset_comp.tex
\begin{table}[ht]
\centering
\caption{
Comparison of our dataset with existing facial-expression datasets containing descriptive text. \textit{Partial} indicates that descriptions mention only coarse cues (e.g., smiling, open mouth) without covering other regions such as the eyes or nose, or conveying the overall emotional context. 
Our dataset offers fine-grained, localized descriptions and additional \textit{implicit} texts describing scenarios likely to elicit the expressions.
}
\begin{tabular}{l c c c c}
\toprule
Dataset       & Samples & Indiv. Expr.&Explicit & Implicit \\ 
\midrule
MMCelebA-HQ & 30k & \cmark&partial & \xmark   \\
CelebAText-HQ     & 15k & \xmark&partial & \xmark   \\
FFHQ-Text  & 760 & \xmark &partial &  \xmark  \\
CelebA-Dialog  &202k   & \cmark &partial & \xmark  \\
\midrule
\DATASET(Ours) & 30k & \cmark & \cmark & \cmark      \\
\bottomrule
\end{tabular}

\label{table:dataset_comp}
\end{table}

%% file: sec/3_method.tex
\section{Method}

In this section, we describe the \OURS framework for generating 3D facial expressions from natural language descriptions. 
Our approach combines a Multimodal Large Language Model (MLLM) with a lightweight expression decoder to establish a robust mapping from text to 3D facial expression parameters. 
We leverage the FLAME model~\cite{FLAME:SiggraphAsia2017} as the face representation, enabling precise control over facial expressions in a structured parameter space. 
In Section~\ref{sec:expression-representation}, we introduce the facial expression parameterization based on FLAME. 
Section~\ref{sec:pipeline-overview} details the overall \OURS pipeline, including the use of the \texttt{\textlangle Expr\textrangle} token and its integration within the MLLM. 
Finally, Section~\ref{sec:training} outlines the multimodal training strategy, including supervision from both text and image data, and the overall optimization objective.

\begin{figure}[ht]
    \centering
    \includegraphics[width=\textwidth]{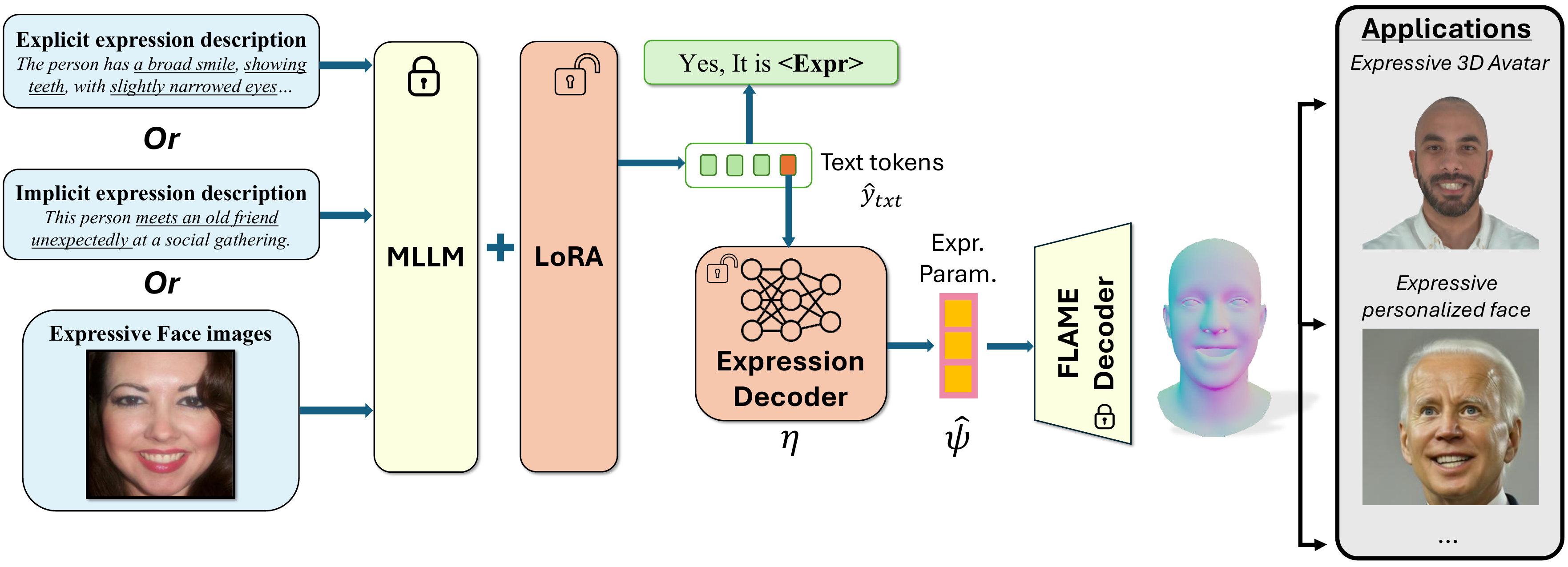}  % Adjust width as needed
    \caption{
    \textbf{Overview of \OURS.} Our framework generates 3D facial expressions from diverse inputs by combining a Multimodal Large Language Model (MLLM) with an expression decoder head $({\eta})$.  \OURS is trained on diverse multimodal supervision, including explicit textual descriptions (e.g., ``a broad smile with narrowed eyes''), implicit contextual prompts (e.g., ``meeting an old friend unexpectedly''), and images.  During training, $\eta$ is optimized, and the LLM within the MLLM is fine-tuned using LoRA while other components remain frozen. 
    The generated expressions can be seamlessly integrated with FLAME-based face avatars for diverse applications.
    }
    \label{fig:ours_architecture}
\end{figure}

\subsection{Facial Expression Representation}
\label{sec:expression-representation}
We adopt the 3D Morphable Model (3DMM) framework~\cite{blanz1999morphable} to represent 3D head geometry with disentangled identity and expression parameters, enabling structured and controllable manipulation.

Specifically, we leverage FLAME~\cite{FLAME:SiggraphAsia2017}, a statistical head model parameterized by identity shape $\beta$, facial expression $\psi$, and pose parameters $\theta$ for rotations around four joints (neck, jaw, and eyeballs), and the global rotation.
FLAME effectively disentangles facial expressions from identity and head pose. In \OURS, we focus on predicting expression parameters ($\psi$) and jaw rotation ($\theta_{jaw}$), while keeping identity ($\beta$) and head pose ($\theta_{head}$) fixed at zero by default. 
This ensures that generated expressions are consistent, disentangled, and placed on an average-sized, neutrally posed head model, i.e., identity-isolated.

\subsection{Pipeline Overview of \OURS}
\label{sec:pipeline-overview}
To establish a robust and generalizable mapping from language to facial expressions, our approach requires a model with strong text understanding capabilities. 
A core design choice in our model is to build on a powerful multimodal large language model (MLLM) backbone. 
Building \OURS on an MLLM enables the model to learn from diverse facial expression data including text-to-3D-face and image-to-3D-face data. 
As a result, our system can efficiently utilize multimodal inputs (as demonstrated in our experimental results in Section~\ref{sec:exp:text2face}).

\textbf{Representing facial expression in language space.} 
To build a connection between the MLLM and facial expression, we treat human facial expressions as a distinct modality and incorporate its representation into the language space. Specifically, we extend the vocabulary of the MLLM to include a new token $\texttt{\textlangle Expr\textrangle}$ that uniquely represents human facial expressions. Given an input text prompt $\mathbf{x}_{txt}$ and/or input image $\mathbf{x}_{img}$, the MLLM $f$ predicts a text response:
\begin{equation}
\mathbf{\hat{y}}_{txt} = f(\mathbf{x}_{img}, \mathbf{x}_{txt}),
\end{equation}
where $\mathbf{\hat{y}}_{txt} =[t_1,\dots,t_N]$ is the output sequence of tokens with corresponding hidden states $[h_1,\dots,h_N]$. When $\mathbf{x}_{txt}$ contains a textual face generation instruction, the predicted response $\mathbf{\hat{y}}_{txt}$ should include a $\langle\texttt{Expr}\rangle$ token, facilitating further 3D face predictions.

\textbf{From \texttt{\textlangle Expr\textrangle} token to 3D facial expression parameters.} 
If one of the output tokens $t_n$ in $\mathbf{\hat{y}_{txt}}$ is the designated \texttt{\textlangle Expr\textrangle} token, the model extracts the hidden state as $h^{\langle\texttt{Expr}\rangle}=h_n \in \mathbb{R}^{4096}$ and projects it using the expression decoder $\eta$ into the latent 3D facial expression parameters $\hat{\psi}=\eta(h^{\langle\texttt{Expr}\rangle}) \in \mathbb{R}^{50}$. 
The predicted expression parameters $\hat{\psi}$ are combined with a prior head template in the FLAME model~\cite{FLAME:SiggraphAsia2017} to produce a 3D head mesh with diverse expressions.
The decoder head $\eta$ is defined as a lightweight MLP for efficiency and simplicity.

\textbf{Combining \OURS with downstream applications.} As \OURS predicts facial expression parameters based on the FLAME head model, it can be combined with many existing FLAME-based methods~\cite{ding2023diffusionrig, qian2024gaussianavatars, zhou2024headstudio} as a conditional input to enable diverse applications in facial avatar generation.
Examples are presented in Figure~\ref{fig:ours_architecture}, where \OURS helps produce expressive 3D avatars and expressive personalized 2D face images.

\subsection{Training}\label{sec:training}
\label{sec:training_details}
\subsubsection{Training Data}
Our training data comprises three primary categories:

\noindent \textbf{Text-to-3D expression data.} 
The text-to-3D expression dataset contains paired text descriptions and facial expression parameters, enabling direct supervision for mapping language to expression parameters.
We employ the following template to guide the MLLMs during training: 
\texttt{"USER: \{description\}, can you give the FLAME expression parameters of this person? ASSISTANT: Sure, it is \textlangle Expr\textrangle."} 
Here, \texttt{\{description\}} denotes a textual expression description and can be replaced with either an explicit or implicit description.

\noindent\textbf{Image-to-3D expression data.} We collect face images from large-scale public face datasets. The face images are processed with a state-of-the-art image-based expressive 3D face reconstruction method~\cite{danvevcek2022emoca} to obtain the corresponding facial expression parameters.
To accommodate the face images into the MLLM training pipeline, we format a template as \texttt{"USER: \textlangle IMAGE\textrangle Can you give the FLAME expression parameters of this person? ASSISTANT: Sure, it is \textlangle Expr\textrangle."}, where $\langle\texttt{IMAGE}\rangle$ represents the face image input. 

\noindent\textbf{Multimodal Instruction-Following Data.}
Following LLaVA-1.5~\cite{liu2023improvedllava}, we include general-purpose VQA data to preserve the model's instruction-following capability.

\subsubsection{Overall Training Objective}
Our overall training objective combines a text-based autoregressive loss with expression reconstruction losses for predicting the ground-truth FLAME expression parameters $\psi$ from either image or text input:
\begin{equation}
\mathcal{L}
=
\lambda_{\mathrm{txt}}\mathrm{CE}(\mathbf{y}_{\mathrm{txt}}, \hat{\mathbf{y}}_{\mathrm{txt}})
+
\lambda_{\mathrm{expr}} |\psi - \hat{\psi}|_2^2
+
\lambda_{\mathrm{mesh}} | \mathbf{w} \odot (M(\psi) - M(\hat{\psi})) |_1 ,
\label{loss_overall}
\end{equation}
where $\mathrm{CE}(\cdot,\cdot)$ denotes the cross-entropy loss between the predicted text tokens $\hat{\mathbf{y}}_{\mathrm{txt}}$ and the ground-truth text tokens $\mathbf{y}_{\mathrm{txt}}$, which helps preserve the model's language understanding and generation capabilities. The expression reconstruction objective consists of two terms: an $\ell_2$ loss on the FLAME expression parameters and an $\ell_1$ geometric loss on the FLAME mesh. Specifically, $M(\cdot)$ maps FLAME expression parameters to the corresponding 3D mesh vertices, and $\mathbf{w}$ is a region-dependent vertex weight map applied element-wise via $\odot$. This weighting emphasizes expression-relevant facial regions while reducing the contribution of less relevant head regions. The coefficients $\lambda_{\mathrm{txt}}$, $\lambda_{\mathrm{expr}}$, and $\lambda_{\mathrm{mesh}}$ balance the contributions of the text, parameter-space, and mesh-space losses, respectively.

%% file: sec/4_experiments.tex
\section{Experiments}
\label{sec:experiments}

\subsection{Experimental Setup}\label{sec:exp:setting}

\textbf{Network Architecture.} Our model is built upon the popular MLLM architecture LLaVA-1.5-7B~\cite{liu2023improvedllava}, using Vicuna-v1.5~\cite{neurips2023vicuna} as the language backbone. To enable efficient fine-tuning, we apply Low-Rank Adaptation (LoRA)~\cite{hu2022lora}. 
On top of the final layer of the MLLM, we attach a lightweight MLP decoder with GeLU activations~\cite{hendrycksG2016gelu} and channel dimensions $[5120, 5120, 50]$. This module maps the multimodal representations to FLAME expression parameters, serving as the expression decoder for 3D facial expression prediction.

\noindent \textbf{Implementation Details.}
We train on 8 NVIDIA A40 GPUs using DeepSpeed~\cite{ras2020deepspeed} with ZeRO~\cite{raj2020zeroopt} optimizer. We use AdamW~\cite{loshchilov2018adamw} with a learning rate of $4\text{e}{-5}$, no weight decay, and a WarmupDecayLR scheduler with 100 warm-up iterations. The loss weights are set to $\lambda_{\mathrm{txt}}=1.0$, $\lambda_{\mathrm{expr}}=1.0$, and $\lambda_{\mathrm{mesh}}=0.01$. The face-region weight $\mathbf{w}$ follows \cite{zielonka2022metricalreconstructionhumanfaces}. We use a per-GPU batch size of 8 and 4 gradient accumulation steps. For compatibility with prior FLAME-based methods, we adopt FLAME 2020 model and represent jaw pose with 6D rotations~\cite{Rot6d_2019_CVPR}.

\noindent \textbf{Datasets.}
As described in Section~\ref{sec:training_details}, \OURS is trained using a diverse set of multi-modal data sources:
(1) \textit{Text-to-3D expression data:} We use both explicit and implicit textual descriptions from our proposed \DATASET dataset as input. The corresponding ground-truth expression parameters are obtained by applying the state-of-the-art expressive face reconstruction method EMOCAv2~\cite{danvevcek2022emoca} to the associated facial images.
(2) \textit{Image-to-3D expression data:} To improve the model's ability to interpret real-world expressions, we include facial images from CelebA~\cite{liu2015celeba}, AffectNet~\cite{affectnet2019}, and FFHQ~\cite{cvpr2019stylegan}, paired with expression parameters extracted using EMOCAv2.
(3) \textit{VQA-style multi-modal data:} To preserve the model’s general multi-modal reasoning capabilities, we incorporate instruction-following data from LLaVA-v1.5-mix665k~\cite{liu2023improvedllava}.

\noindent \textbf{Evaluation Metrics.}
We evaluate both explicit and implicit text-to-3D facial expression synthesis. 
Following standard practice in expressive 3D face reconstruction~\cite{danvevcek2022emoca, retsinas20243dfacialexpressionsanalysisbyneuralsynthesis, toisoul2021emonet}, we assess semantic expression fidelity using an emotion recognition network on generated 3D face meshes. 
This network predicts three types of emotion-related signals from the 3D face mesh: (1) \textit{valence}: positivity or negativity of the expression, (2) \textit{arousal}: intensity of the emotion, and (3) the discrete emotion category. 
We report Concordance Correlation Coefficient (CCC) for valence and arousal, and top-1 accuracy for emotion classification, using AffectNet annotations as ground truth~\cite{affectnet2019}. 
For methods that predict FLAME parameters, we further report the L1 distance between predicted and ground-truth 3D point clouds over the head, face and lip regions. Additional metric details are provided in the supplementary.

\input{tables/description_benc_full}

\begin{figure}[ht]
  \centering
   \includegraphics[width=\linewidth]{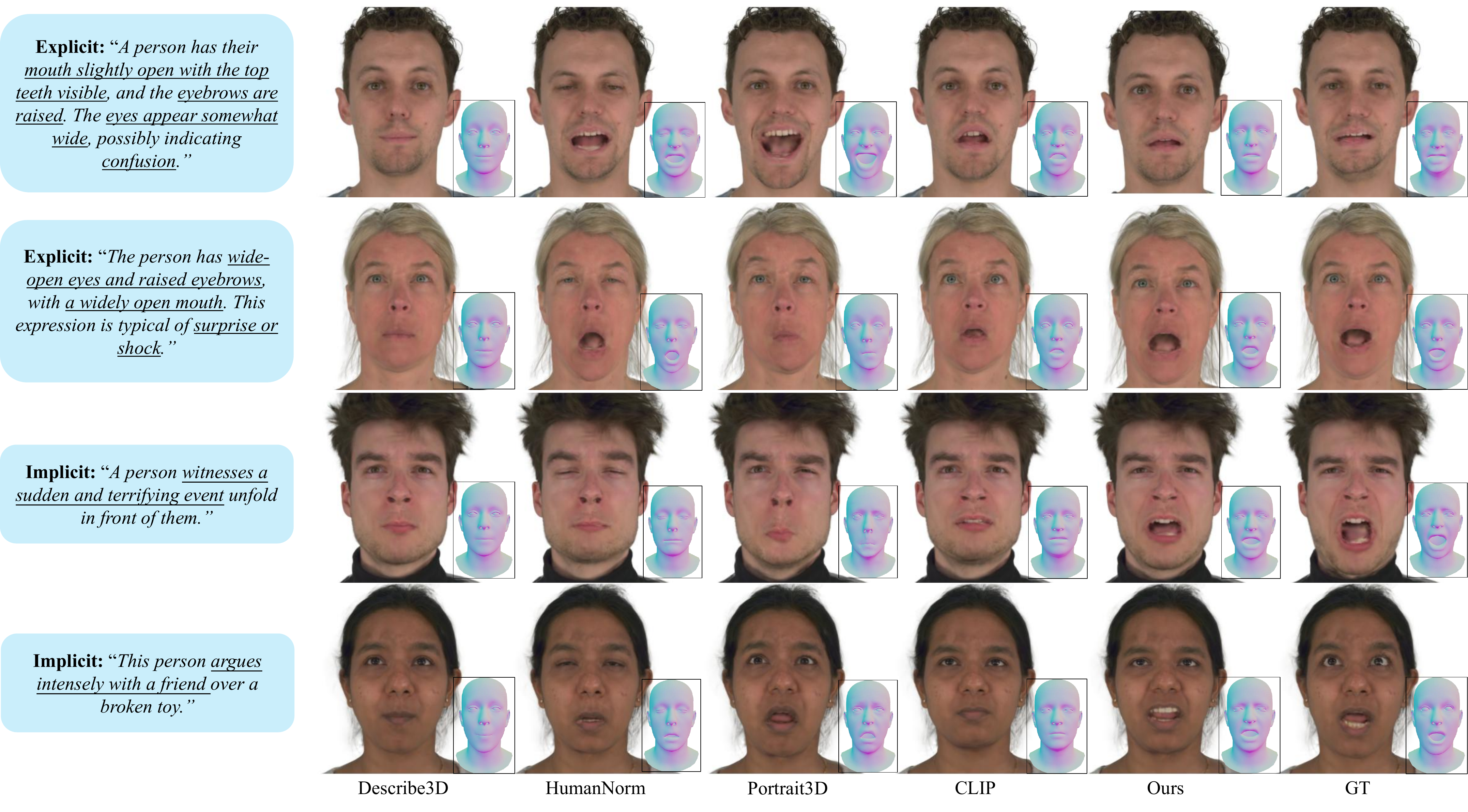}

   \caption{
   Qualitative comparison of \OURS with relevant methods. Results are visualized in both the 3DMM space and rendered 3D avatars using GaussianAvatar. Existing models often struggle with complex explicit descriptions and even simple implicit ones. In contrast, \OURS produces semantically faithful and expressive results across diverse inputs. Please zoom in for more details.
   }
   \label{fig:vis_res}
\end{figure}

We evaluate the effectiveness of \OURS in generating 3D facial expressions from natural language prompts. Section~\ref{sec:exp:setting} outlines the experimental setup, including model configurations, datasets, and evaluation metrics. In Section~\ref{sec:exp:text2face}, we provide quantitative and qualitative comparisons with state-of-the-art methods on both explicit and implicit facial expression generation. Section~\ref{sec:exp:ablation} presents an ablation study that examines the impact of different training data modalities. Finally, Section~\ref{sec:exp:avatar} demonstrates downstream applications of our method in photorealistic avatar generation and personalized face synthesis.

\subsection{Text to 3D Facial Expression Generation}\label{sec:exp:text2face}

We evaluate our method’s ability to generate 3D facial expressions from diverse textual descriptions by comparing it with baselines from two related paradigms: (1) text-to-expression parameter regression methods, which are most closely aligned with our task setting, and (2) recent text-to-3D head generation methods, which also produce expressive 3D faces. Although our approach follows the text-to-expression regression paradigm, 
we include the latter category to provide a broader comparison.

First, since our method operates within a text-to-expression parameter regression framework, we compare it with baselines in this setting. These include an early text-to-3DMM generation method Describe3D~\cite{describe3d2023}, as well as two additional models we design: a CLIP-based regressor and a unimodal LLM-based regressor. These two regressors are trained using the same pipeline as our model to ensure fair and controlled comparison.

We also compare \OURS with state-of-the-art text-to-3D head methods synthesizing full head geometry from text, including HumanNorm~\cite{huang2023humannorm} and Portrait3D~\cite{Portrait3D_sig24}. As these methods typically rely on computationally intensive iterative refinement, large-scale quantitative evaluation is impractical. Instead, we provide qualitative comparisons and conduct a perceptual user study against Portrait3D, the strongest representative in this category.

\noindent\textbf{Quantitative comparison} 
We provide the emotion evaluation results and geometric errors for quantitative comparison on the explicit and implicit-based 3D facial expression synthesis benchmark. 
To contextualize the difficulty of the benchmark, we also include a static FLAME head with neutral expressions as a baseline.
As shown in Table~\ref{table:bench_res}, the CLIP baseline trained on our \DATASET dataset, already surpasses Describe3D in expression generation. Notably, \OURS consistently outperforms both the CLIP and LLM(Vicuna)-based baselines across both explicit and implicit tasks, demonstrating its superior ability to capture fine-grained and context-dependent expressions. 
The relatively low performance of the FLAME baseline further indicates that our benchmark comprises diverse and challenging expressions.

Interestingly, Table~\ref{table:bench_res} also highlights the strong generalization ability of our MLLM-based approach. Even when trained without implicit descriptions, \OURS generalizes well to unseen implicit inputs, while maintaining high performance on explicit expressions. This suggests that our architecture and training strategy enable robust expression modeling beyond the training distribution.

We also try to compare \OURS with diffusion-based text-to-3D head generation methods quantitatively. 
Since these methods are generally time-consuming to synthesize a static head with a single query, it would be impractical to perform large-scale comparison, 
% Since these methods require costly iterative optimization for synthesizing each static head, large-scale evaluation is impractical.
We therefore sample 50 explicit and 50 implicit descriptions from our benchmark with balanced emotion distributions, and fit FLAME to each generated head using an image-based face tracker~\cite{danvevcek2022emoca}.

Table~\ref{table:subset_comp} presents the comparison results. Portrait3D performs better than HumanNorm on explicit texts but they perform similarly poorly for implicit texts, while \OURS consistently achieves better facial expression synthesis quality than other diffusion-based text-to-3D head methods.

\noindent\textbf{Qualitative comparison.} 
We qualitatively compare \OURS with state-of-the-art text-to-3D face generation methods, including HumanNorm~\cite{huang2023humannorm} and Portrait3D~\cite{Portrait3D_sig24}. Since these methods jointly synthesize identity, expression, and texture from text, we fit the FLAME model to their outputs to isolate expression quality, as shown in Figure~~\ref{fig:vis_res}.

We observe that all models perform better on explicit prompts, which contain clearer emotion signals, while implicit prompts remain more challenging.
HumanNorm, trained with depth- and normal-adapted diffusion on full-body datasets, struggles to interpret facial expressions from natural language inputs. Portrait3D generalizes better due to its strong 3D-aware GAN prior, but still fails to capture fine-grained expression detail—particularly under complex, nuanced descriptions. 
In contrast, both the CLIP-based baseline and \OURS produce more semantically aligned expressions, 
with \OURS showing clear advantages in subtle and abstract cases.

\begin{table}[t]
\centering
\begin{minipage}[t]{0.48\linewidth}
\centering
\captionof{table}{Quantitative comparisons to state-of-the-art text-to-3D face generation methods. \OURS obtains the best expression synthesis quality compared to other text-driven methods. }
\resizebox{0.9\linewidth}{!}{
\begin{tabular}{c|l|ccc}
\toprule
Eval & Method & $L1_{Head}\downarrow$ & $L1_{Face}\downarrow$ & $L1_{Lip}\downarrow$\\
\midrule
\multirow{3}{*}{Exp}
& HumanNorm & 1.03 & 1.75&4.36\\
& Portrait3D & 0.92 & 1.51 &3.64\\
& \OURS & \textbf{0.62} & \textbf{1.01}&\textbf{2.37}\\
\midrule
\multirow{3}{*}{Imp}
& HumanNorm & 1.05 & 1.80&4.64\\
& Portrait3D & 1.03 & 1.74&4.43 \\
& \OURS & \textbf{0.70} & \textbf{1.19} &\textbf{2.93}\\
\bottomrule
\end{tabular}
}
\label{table:subset_comp}

\end{minipage}
\hfill
\begin{minipage}[t]{0.48\linewidth}
\centering
\captionof{table}{User study results comparing \OURS and Portrait3D on explicit and implicit text prompts. ‘++’ indicates strong preference, ‘+’ indicates weak preference. \OURS is preferred in 75\% of cases overall, while Portrait3D is preferred in only 14\%.}
\resizebox{\linewidth}{!}{
\begin{tabular}{l|cc|c|cc}
\toprule
 & \multicolumn{2}{|c|}{\textbf{\OURS}} & Indifferent &\multicolumn{2}{|c}{\textbf{Portrait3D}}\\
Prompt Type & \textbf{++} & \textbf{+} & \textbf{} & \textbf{+} & \textbf{++} \\
\midrule
Explicit   & 42\% & 33\% & 7\%  & 6\%  & 10\% \\
Implicit   & 46\% & 30\% & 10\% & 10\% & 2\%  \\
Overall    & 44\% & 31\% & 8\%  & 8\%  & 6\%  \\
\bottomrule
\end{tabular}
}
\label{table:user_study_transposed}
\end{minipage}
\end{table}

\noindent\textbf{Perceptual user study.} 
Evaluating the semantic alignment between textual descriptions and 3D facial expressions remains a challenging task. To assess perceptual quality, we conducted a user study with 25 participants, all with backgrounds in computer graphics.
Each participant was shown 20 text prompts along with rendered expressions from \OURS and Portrait3D. Since Portrait3D outputs full head geometry, we fitted FLAME parameters to its results for a fair comparison. Participants selected the rendering that best matched each prompt. 
As summarized in Table~\ref{table:user_study_transposed}, \OURS was preferred in 76\% of cases, while Portrait3D was selected in only 14\%, with the remaining 10\% indicating no clear preference. These results suggest that \OURS produces facial expressions that more accurately reflect both explicit and implicit textual cues.

\input{tables/ablation_training_data}

\noindent\subsection{Ablation experiments}\label{sec:exp:ablation}
We conduct ablations to evaluate the contributions of different training modalities and the design of the expression grounding token.

\noindent \textbf{Influence of training data}
As shown in Table~\ref{table:ablation_data} and visual examples in supplementary, models trained only on paired text-to-expression data perform worse than those trained with additional multimodal supervision. Incorporating general-purpose VQA data and face image–based expression labels consistently improves performance across input types, demonstrating the effectiveness of our multimodal training strategy.

\noindent \textbf{Design of \texttt{\textlangle Expr\textrangle} token}
Table~\ref{tab:expr_token_comp} compares our dedicated \texttt{\textlangle Expr\textrangle} token with two alternatives: mean pooling over text hidden states and using the \texttt{\textlangle EOS\textrangle} token as the expression representation. Under the same setting, \texttt{\textlangle Expr\textrangle} achieves the best performance across both explicit and implicit inputs. Mean pooling dilutes expression-relevant cues with unrelated textual information, while \texttt{\textlangle EOS\textrangle} entangles expression grounding with sequence termination.

\begin{table}[t]
\centering
\caption{Comparison of our \texttt{\textlangle Expr\textrangle} design with other alternative designs.}
\begin{tabular}{l|ccc|ccc}
\toprule
\multirow{2}{*}{Method} 
& \multicolumn{3}{c|}{Explicit} 
& \multicolumn{3}{c}{Implicit} \\
& $L1_{Head}\downarrow$ & $L1_{Face}\downarrow$ & $L1_{Lip}\downarrow$
& $L1_{Head}\downarrow$ & $L1_{Face}\downarrow$ & $L1_{Lip}\downarrow$\\
\midrule
Mean Pooling 
& 0.62 & 1.00 & 2.35 
& 0.68 & 1.15 & 2.85 \\
\texttt{<EOS>} Token 
& 0.75 & 1.21 &3.05 
& 0.84 & 1.43 &3.93 \\
\texttt{<Expr>} Token 
& \textbf{0.60} & \textbf{0.98}&\textbf{2.29}
& \textbf{0.66} & \textbf{1.09} &\textbf{2.69} \\
\bottomrule
\end{tabular}
\label{tab:expr_token_comp}
\end{table}

\subsection{Further applications}
\label{sec:exp:avatar}

\OURS predicts facial expression within the FLAME model space, making it directly compatible with existing FLAME-based avatar systems. Beyond qualitative comparisons in Figure~\ref{fig:vis_res}, here we show two downstream applications.

\begin{figure}
    \centering
    \includegraphics[width=\linewidth]{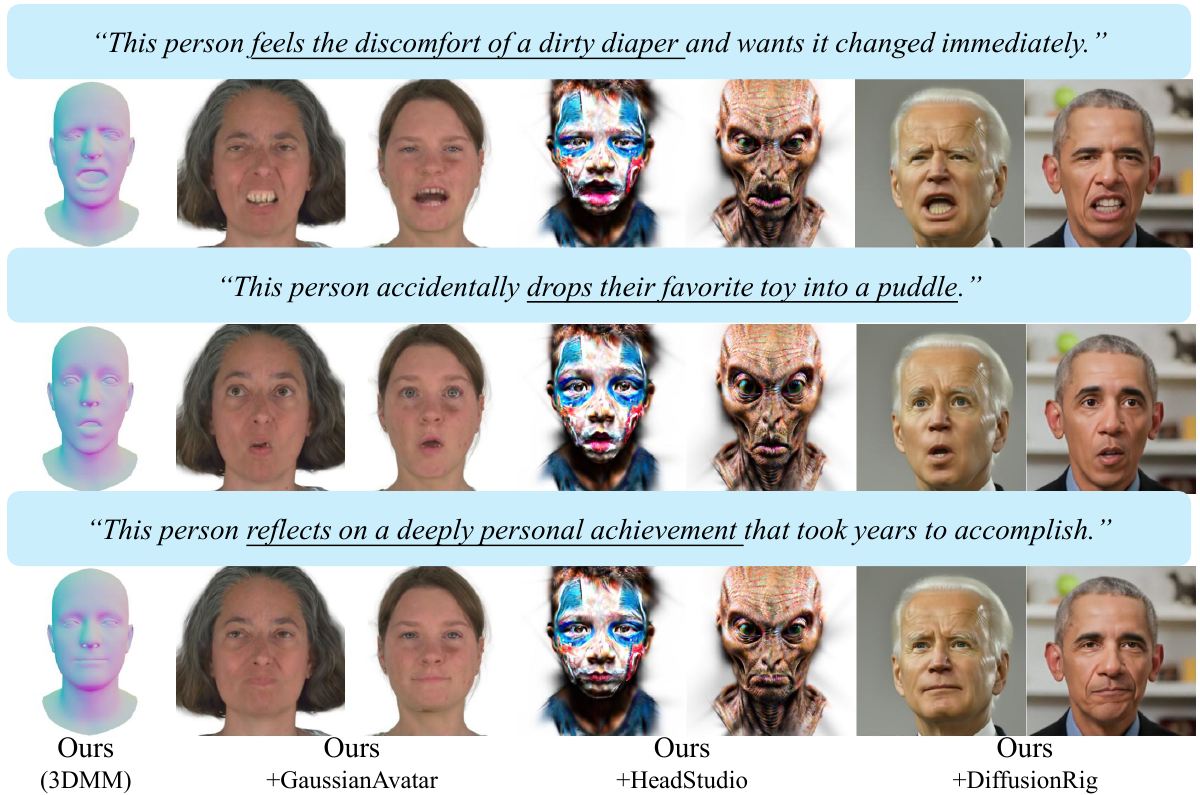}
    \caption{Diverse applications of \OURS. Combining \OURS with GaussianAvatar, HeadStudio and DiffusionRig enables downstream tasks such as personalized expressive 3D avatar generation and 2D face image synthesis.}
    \label{fig:applications}
\end{figure}

\begin{itemize}

\item \textbf{Expressive 3D avatar generation.}
\OURS enables expressive control of 3D avatars through text, complementing existing 3D identity and texture synthesis pipelines like GaussianAvatar~\cite{qian2024gaussianavatars} and HeadStudio \cite{zhou2024headstudio}.

\item \textbf{Expressive 3D-aware personalized image generation.} By combining \OURS with DiffusionRig~\cite{ding2023diffusionrig}, our method can replace the expression to synthesize identity-preserving, expression-aware face images.
\end{itemize}

Figure~\ref{fig:applications} demonstrates the versatility of \OURS across diverse pipelines, with additional results provided in the supplementary.

%% file: tables/description_benc_full.tex
\begin{table}[ht]
\centering
\caption{Quantitative results for text-to-3D facial expression generation with different descriptions. The best performance is highlighted with \textbf{bold} font. \OURS outperforms other baselines across explicit and implicit benchmarks. \OURS also shows the best generalization ability even when trained without implicit descriptions, maintaining strong performance on explicit and unseen inputs.
% \TODO{put a x mark in table}
}
\resizebox{\textwidth}{!}{
\begin{tabular}{l|cc|cccccc|cccccc}
\toprule
&\multicolumn{2}{c|}{Training} & \multicolumn{6}{c|}{Explicit Expression Descriptions} &\multicolumn{6}{c}{Implicit Expression Descriptions}\\
Method       &Exp & Imp & V-CCC $\uparrow$& A-CCC $\uparrow$ & E-ACC$\uparrow$ &$L1_{Head} \downarrow$ & $L1_{Face}\downarrow$ & $L1_{Lip} \downarrow$& V-CCC $\uparrow$ & A-CCC $\uparrow$ & E-ACC$\uparrow$ & $L1_{Head}\downarrow$ & $L1_{Face}\downarrow$ & $L1_{Lip}\downarrow$\\ 

\midrule
FLAME &  &  & -    & -& -& 0.83 & 1.41 &3.53& -    & -& -& 0.83 & 1.41 & 3.57\\
\midrule
Describe3D     & & & 0.48    & 0.34  &0.34 &-&-&0.19     & 0.08  & 0.19 &-&- \\
\midrule
CLIP              & \checkmark &&  0.61    & \textbf{0.57} & 0.51 & 0.66& 1.06 & 2.46 &0.36     & 0.28  & 0.32 &0.84& 1.45 & 3.70\\
Vicuna-1.5-7b     &\checkmark & &  0.56     & 0.53 & 0.51 &0.66& 1.08&2.48&  0.47    & 0.34  & 0.38 & 0.80 & 1.35& 3.37\\
\OURS             &\checkmark & &  \textbf{0.61}   & 0.50  & 0.51 & \textbf{0.61}&\textbf{1.00} & \textbf{2.36}&  \textbf{0.49}    & \textbf{0.44}  & \textbf{0.41} & \textbf{0.74}& \textbf{1.30}&\textbf{3.27}\\
\midrule
CLIP     &\checkmark & \checkmark&  0.64    & 0.56   & 0.52 &0.65& 1.07 &2.50& 0.49    & 0.41  & 0.41 &0.73 &1.24 & 3.04\\
Vicuna-1.5-7b                &\checkmark &\checkmark & 0.60     & 0.55  & 0.52 &0.65& 1.08 & 2.50& 0.49     & 0.39  & 0.41 & 0.71& 1.21&2.94\\
\OURS             &\checkmark &\checkmark &\textbf{0.68}  &  \textbf{0.62}  & \textbf{0.59}& \textbf{0.60}& \textbf{0.98} &\textbf{2.29} &  \textbf{0.57}    & \textbf{0.48}  &\textbf{0.44} &\textbf{0.66}& \textbf{1.09}& \textbf{2.69}\\
\bottomrule
\end{tabular}
}

\label{table:bench_res}
\end{table}

%% file: tables/ablation_training_data.tex
\begin{table}[ht]
\centering
\caption{Ablation study illustrating the impact of different training data modalities on text-to-3D facial expression synthesis. 
% \OURS achieves strong performance on explicit descriptions even when trained solely on text-to-expression pairs. 
Incorporating facial images data improves \OURS performance on implicit descriptions, demonstrating the effectiveness of multi-modal supervision for enhancing generalization.
}
\renewcommand{\arraystretch}{0.95}
\setlength{\tabcolsep}{2pt} 
\resizebox{0.9\textwidth}{!}{
\begin{tabular}{ccc|cc|cccccc}
\toprule
% Method       & Unsupervised &CD $\downarrow$ & CR $(\%)\uparrow$  & RFRR $\uparrow$ & User Study $(\%)\uparrow$\\ 
\multicolumn{3}{c|}{Training} & \multicolumn{2}{c|}{Eval} &&&&&\\

Text&VQA&Image& Exp& Imp& V-CCC $\uparrow$ & A-CCC $\uparrow$  & E-ACC$\uparrow$ & $L1_{Head}\downarrow$ & $L1_{Face}\downarrow$ & $L1_{Lip}\downarrow$\\ 
% \hline 
\midrule
\checkmark     &  &   &  \checkmark  & & 0.58     & 0.56  & 0.52 & 0.64& 1.06 &2.49\\
\checkmark     & \checkmark &   & \checkmark   & & 0.66    & 0.61 & 0.56 & \textbf{0.60}& 1.02 &2.32\\
\checkmark    & \checkmark & \checkmark  & \checkmark   & & \textbf{0.68}    & \textbf{0.62}  & \textbf{0.59} & \textbf{0.60}& \textbf{0.98} &\textbf{2.29}\\
\midrule
\checkmark     &   &   &    & \checkmark& 0.55   & 0.37    & 0.41 & 0.71& 1.21 &2.98\\
\checkmark     & \checkmark &   &   &\checkmark & 0.55    & 0.46  &0.42 & 0.67& 1.15&2.73\\
\checkmark    & \checkmark & \checkmark  &    & \checkmark& \textbf{0.57}    & \textbf{0.48}  & \textbf{0.44} &\textbf{0.66}& \textbf{1.09}&\textbf{2.69}\\

\bottomrule
\end{tabular}
}
% \vspace{-.5cm}
\label{table:ablation_data}
\end{table}

%% file: sec/5_limitations.tex
\section{Limitations}
\label{sec:limitations}
Our study adopts FLAME for 3D facial expressions. Although FLAME is widely used and provides a standardized, relatively identity-isolated expression space, it has limited expressiveness. Its training data does not fully cover the diversity of human faces across ages, ethnicities, and cultures, and it may miss subtle geometric details such as wrinkles and fine-grained muscle movements. Moreover, FLAME does not achieve complete identity--expression disentanglement~\cite{IdExpAmbiguityFlame2021}. Therefore, our predicted FLAME parameters should be viewed as an intermediate expression control signal rather than a final high-fidelity facial representation. Such signals can be instantiated by downstream renderers or avatar-generation methods, and our modular framework can be extended to richer blendshape models, neural facial representations, or future 3D morphable models.

Our ground-truth 3D expressions are obtained via a state-of-the-art monocular 3D face estimation method. Although this setup may still struggle under low-quality inputs or extreme poses, it offers a scalable and robust solution for constructing large-scale paired text-to-3D datasets. We employ AffectNet and EMOCAv2 for their diversity and robustness, and anticipate continued improvements in 3D expression estimation to further improve our setting.

Finally, our textual supervision relies on captions generated by GPT-4o. While these captions may reflect limited cultural or contextual coverage, GPT-4o remains a strong vision--language model for this task. Ongoing advances in large language models are expected to enable richer, fairer, and more inclusive descriptions of facial expressions.

%% file: sec/5_conclusion.tex
\section{Conclusion}
\label{sec:conclusion}

We introduced \OURS, a framework for generating 3D facial expressions from natural language using an MLLM with a dedicated \texttt{\textlangle Expr\textrangle} token and a lightweight decoder.
To support this task, we introduced \DATASET, a dataset of 3D facial expressions paired with rich textual annotations, including explicit appearance-based captions and implicit contextual cues. 
On the \DATASET benchmark, \OURS produces higher-quality expressions than existing text-to-3D head and expression methods in emotion accuracy, expression realism, and generalization. Experiments and user studies further validate its ability to generate fine-grained 3D facial expressions, and we demonstrate its versatility in 3D avatar generation and personalized 2D editing. Overall, this work advances intuitive, text-driven expressive avatar creation and establishes a foundation for broader exploration of MLLMs in 3D human-centric generation.

%% file: sec/X_suppl.tex
\clearpage
 {
   \newpage
        \centering
        \Large
        % \textbf{\papertitle}\\
        \vspace{0.5em}Supplementary Material \\
        \vspace{1.5em}
   }

\appendix

\section{Details of data generation}
\label{sec:data_generation}
\begin{figure}[ht]
  \centering
   \includegraphics[width=\linewidth]{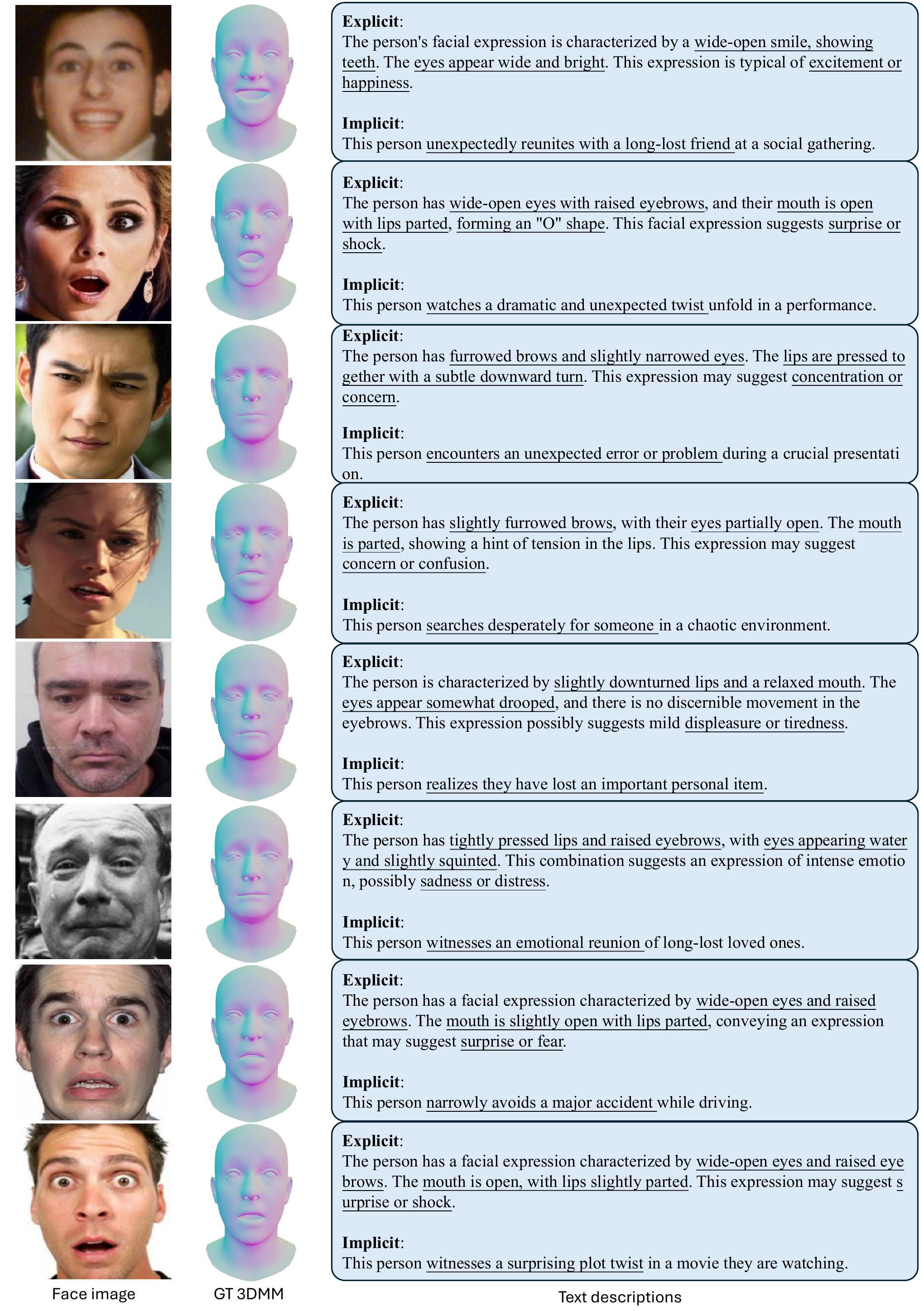}

   \caption{Examples of paired data in \DATASET}
   \label{fig:supp:dataset_samples}
\end{figure}

In this section, we describe the data generation procedure. Both the training and evaluation sets are produced using the automatic pipeline described below, while the test set is further verified through additional human inspection.
Representative annotated samples from our dataset are shown in Figure~\ref{fig:supp:dataset_samples}.

\textbf{Image-to-3D expression data}
We first apply a face detector~\cite{bulat2017far} to the face image dataset to detect 2D facial landmarks, and use the detected landmarks to crop and align the face images. Noisy samples are discarded during this preprocessing stage. 
We then remove duplicate images from the filtered dataset by computing CLIP embeddings and applying a cosine-similarity threshold of 0.93. 
For each remaining aligned face image, we run EMOCAv2 to obtain a FLAME reconstruction, and use the estimated FLAME expression and jaw parameters as the corresponding 3D expression representation. 
To filter inaccurate reconstructions, we compute the mean L1 distance between the detected 68 2D facial landmarks and the projected 68 landmarks from the estimated FLAME mesh. Samples are discarded if the mean L1 distance exceeds 17.9 over the full face or 1.25 over the lip region. 
The final filtered set of face images, together with their corresponding FLAME expression and jaw parameters, is used for training.

\begin{figure}[ht]
  \centering
   \includegraphics[width=\linewidth]{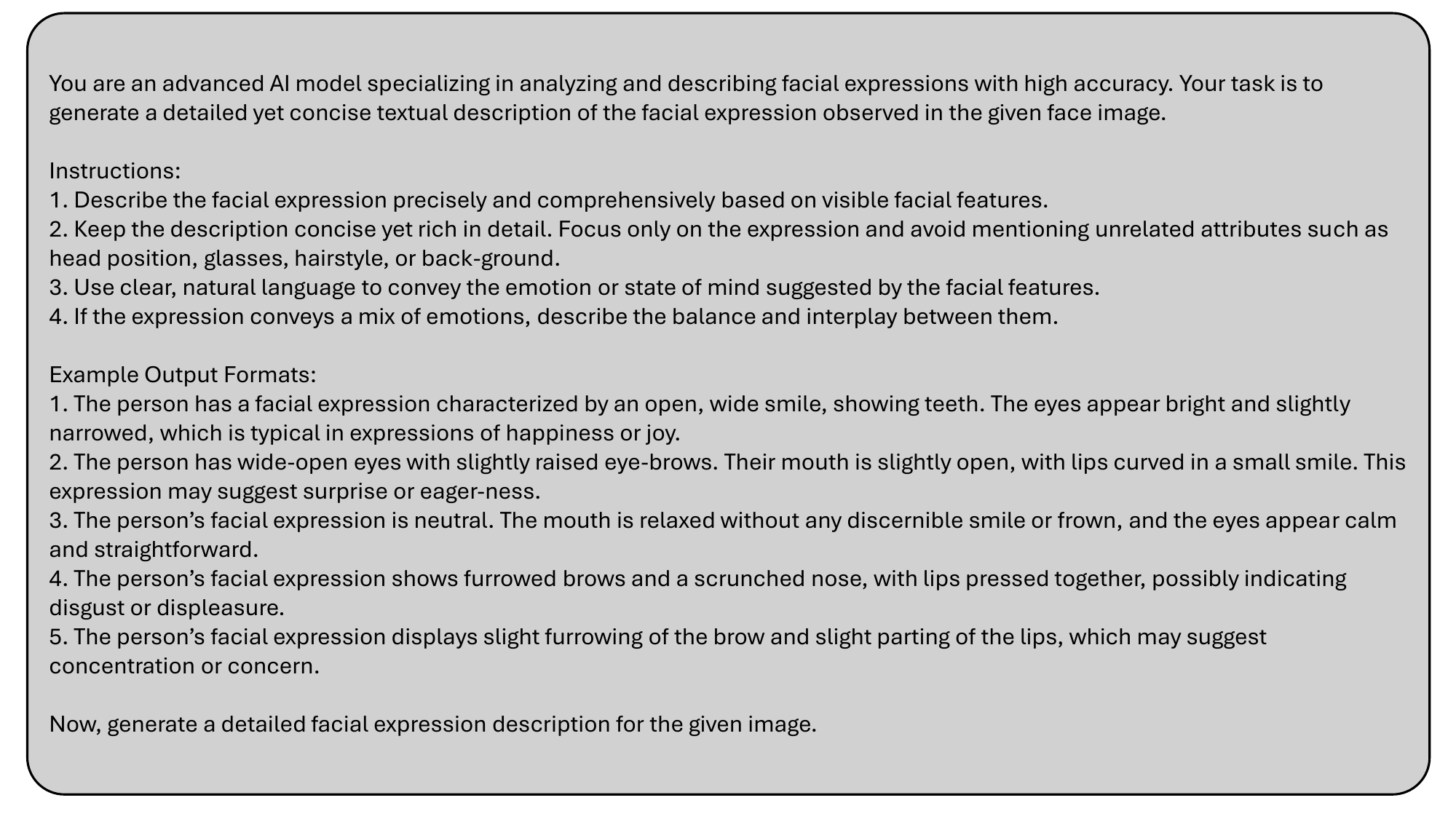}

   \caption{Prompts for querying GPT4o to generate explicit facial expression descriptions on face images.}
   \label{fig:supp:gpt_prompt1}
   \vspace{-.15cm}
\end{figure}

\begin{figure}[ht]
  \centering
   \includegraphics[width=\linewidth]{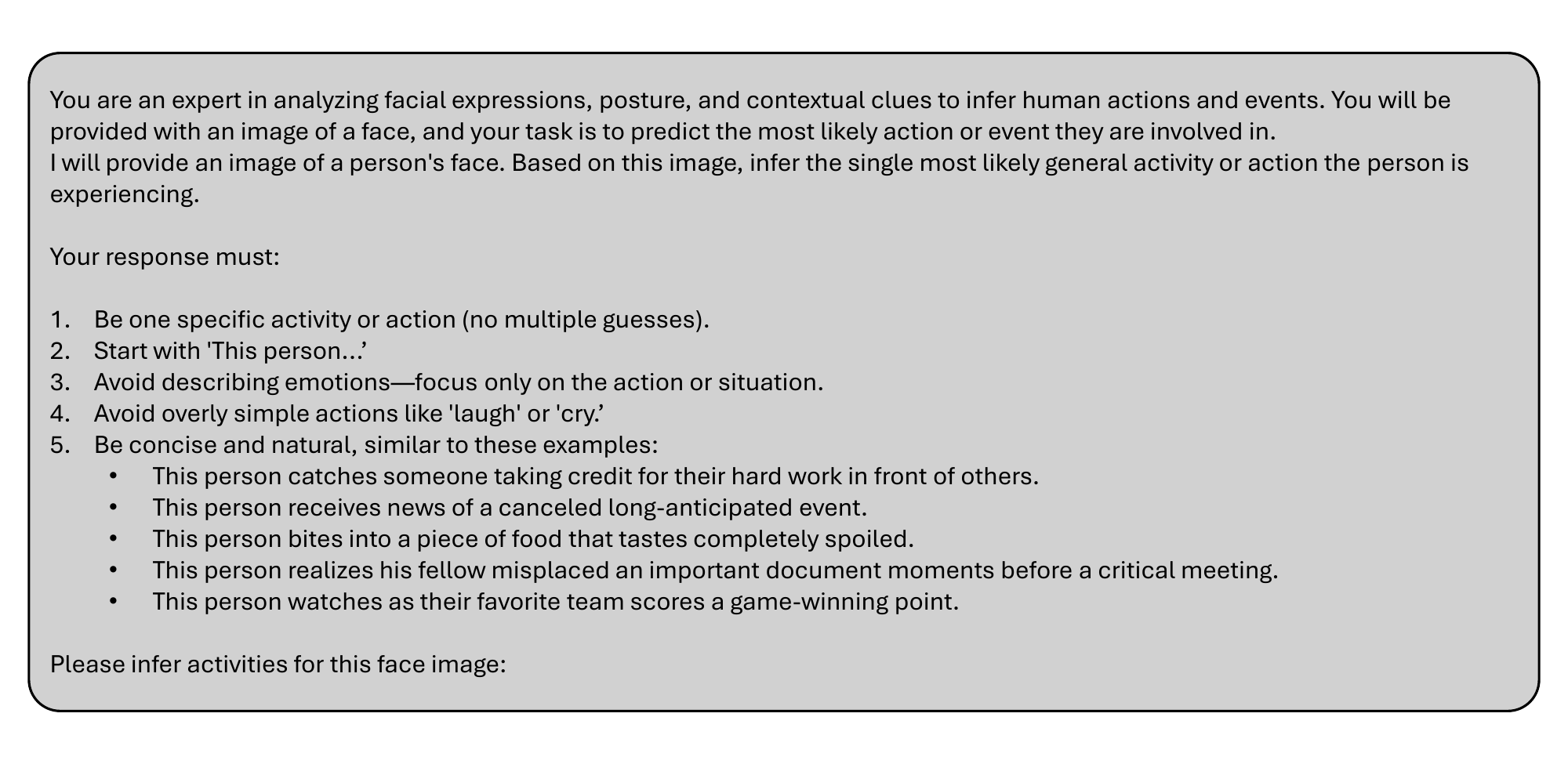}

   \caption{Prompts for querying GPT4o to generate implicit facial expression descriptions on face images}
   \label{fig:supp:gpt_prompt2}
   \vspace{-.15cm}
\end{figure}

\textbf{Text-to-3D expression data}
We follow the same filtering procedure as in the image-to-3D expression data generation stage to remove noisy face images and unreliable expression representations. We then subsample the images uniformly according to their emotion annotations and query GPT-4o to generate both explicit and implicit textual descriptions for each image, using the prompts shown in Fig.~\ref{fig:supp:gpt_prompt1} and Fig.~\ref{fig:supp:gpt_prompt2}, respectively. To reduce redundancy and hallucination, we deduplicate the GPT-4o outputs using the MinHash LSH algorithm~\cite{lsh1998} with a threshold of 0.9. Following EmoNet~\cite{toisoul2021emonet}, we retain test samples whose emotion labels are consistent with the corresponding valence and arousal annotations.

\section{Evaluation protocol}
\textbf{Emotion recognition metrics}:
In the main paper, we evaluate the performance on valence and arousal in the same setting as described in ~\cite{toisoul2021emonet, danvevcek2022emoca}:

The computation of Concordance correlation coefficient (CCC) requires to compute Pearson correlation coefficient (PCC), which measures the correlation between $Y$ and  $\hat{Y}$:
\begin{equation}
        PCC(Y, {\hat{Y}}) = \frac{E(Y-{\mu_Y})(Y-\mu_{\hat{Y}})}{{\sigma_Y}{\sigma_{\hat{Y}}}} ,
\end{equation}

CCC consists of the PCC with a penalty to the correlated signals with different means:
\begin{equation}
        CCC(Y, {\hat{Y}}) = \frac{2{\sigma_Y}{\sigma_{\hat{Y}}}PCC(Y, {\hat{Y}})}{{\sigma_Y}^2+{\sigma_{\hat{Y}}}^2 + {(\mu_Y-\mu_{\hat{Y}})}^2} ,
\end{equation}

\noindent\textbf{Emotion Recognition Network}: To evaluate the emotion recognition metrics, we have to fit an individual emotion recognition network for the predictions from different methods. Following \cite{danvevcek2022emoca}, we fit a 4-layer MLP with batch normalization for regressing valence and arousal levels and classifying emotion categories. The emotion recognition network would take the predicted output 3DMM parameters from each text-to-3D face generation method as input.
For each emotion recognition network, we train for 50 epoches and report the results with the last checkpoint.

\noindent\textbf{Optimization of Emotion Recognition Network}: We utilize the same overall losses as described in \cite{toisoul2021emonet} for optimization:

\begin{equation}
\begin{split}
        L(Y, {\hat{Y}}) = CE(Y, {\hat{Y}})+\frac{\alpha}{\alpha+\beta+\gamma}L_{MSE}(Y, {\hat{Y}}) \\
        + \frac{\beta}{\alpha+\beta+\gamma}L_{PCC}(Y, {\hat{Y}})+\frac{\gamma}{\alpha+\beta+\gamma}L_{CCC}(Y, {\hat{Y}}),
    \end{split}
\end{equation}
where $\alpha$, $\beta$ and $\gamma$ are shake-shake regularization coefficients uniformly sampled from the interval $[0,1]$ for each training batch.
$CE(Y, {\hat{Y}}$ refers to emotion classification loss between predicted emotion category and the ground truth category. And:

\begin{equation}
        L_{MSE}(Y, {\hat{Y}}) = MSE_{valence}(Y,{\hat{Y}})+MSE_{arousal}(Y,{\hat{Y}}) ,
\end{equation}

\begin{equation}
        L_{PCC}(Y, {\hat{Y}}) = 1-\frac{PCC_{valence}(Y,{\hat{Y}})+PCC_{arousal}(Y,{\hat{Y}})}{2} ,
\end{equation}

\begin{equation}
        L_{CCC}(Y, {\hat{Y}}) = 1-\frac{CCC_{valence}(Y,{\hat{Y}})+CCC_{arousal}(Y,{\hat{Y}})}{2} ,
\end{equation}

The training batch size for emotion network is 64. An Adam optimizer with a learning rate of 0.0001, $\beta_1=0.9$ and $\beta_2=0.999$

\begin{figure}[ht]
  \centering
   \includegraphics[width=\linewidth]{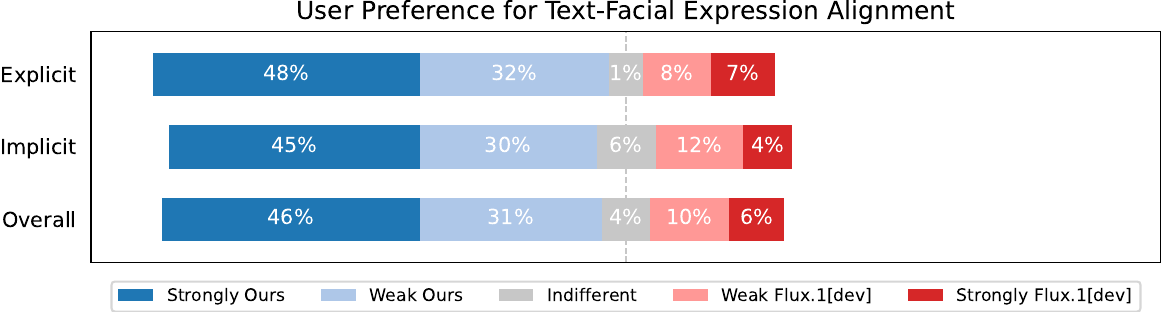}

   \caption{
   Perceptual study comparing \OURS and Flux.1[dev]. 
   Participants preferred \OURS in both explicit and implicit input settings. overall, \OURS was selected in 77\% of cases, compared with 16\% for Flux.
   }
   \label{fig:user_study_flux}
\end{figure}

\section{Comparison with text-based 2D generative models}
Recent progress in text-to-image generation has greatly improved image quality and text alignment. A natural baseline for text-driven 3D facial expression generation is to first synthesize a 2D expressive face image from text and then fit a 3DMM to the generated image.

To evaluate this alternative, we compare \OURS with three strong text-to-image baselines, SDXL~\cite{podell2024sdxl}, Flux.1[dev]~\cite{flux2024} and Z-Image-Turbo~\cite{team2025zimage}. For a fair comparison in 3D expression space, we fit FLAME models to the images generated by these methods using EMOCAv2~\cite{danvevcek2022emoca}.

Table~\ref{table:flux_comp} reports the quantitative comparison. 
\OURS consistently outperforms SDXL, Flux.1[dev] and Z-Image-Turbo in facial expression quality across all metrics, highlighting the advantage of directly predicting 3D facial expressions from text instead of relying on 2D image synthesis followed by 3D fitting. 
Figure~\ref{fig:user_study_flux} shows the perceptual comparison between Flux.1[dev] and \OURS. 
Participants preferred \OURS under both explicit and implicit input settings. 
Overall, \OURS was selected in 77\% of cases, whereas Flux was chosen in only 16\%, indicating that \OURS produces facial expressions that more faithfully match the textual descriptions.

\begin{table}[t]
\centering
\caption{Quantitative comparison with text-to-image generative model. Exp stands for Explicit descriptions and Imp represents Implicit descriptions.}
\begin{tabular}{c|l|ccc}
\toprule
Eval & Method & $L1_{Head}\downarrow$ & $L1_{Face}\downarrow$ & $L1_{Lip}\downarrow$\\
\midrule
\multirow{3}{*}{Exp}
& SDXL & 1.27 & 2.10 & 5.38\\
& Flux.1[dev] & 1.13 & 1.92 & 4.88\\
& Z-image-Turbo & 0.89 & 1.50 & 3.57\\
& \OURS & \textbf{0.60} & \textbf{0.98} &\textbf{2.29}\\
\midrule
\multirow{3}{*}{Imp}
& SDXL & 1.34 & 2.22 & 5.33\\
& Flux.1[dev] & 1.43 & 2.33 & 5.50\\
& Z-image-Turbo & 1.26 & 2.12 & 5.22\\
& \OURS & \textbf{0.66} & \textbf{1.09} &\textbf{2.69}\\
\bottomrule
\end{tabular}
\label{table:flux_comp}
\end{table}

\section{Ablation on loss function}
In Eq.2 of main paper, our training objective consists of a cross-entropy loss for text generation, together with an $\ell_2$ FLAME parameter loss and a weighted $\ell_1$ geometry loss for facial expression prediction. 
To study the effect of the expression prediction losses, we compare our facial expression objective with two variants that use only the FLAME loss or only the geometry loss, while keeping all other training settings unchanged. 

As shown in Table~\ref{tab:ablation_loss}, using either loss alone leads to inferior performance. 
The FLAME loss provides direct supervision in the parametric expression space, while the geometry loss constrains the resulting facial surface, making them complementary. 
Combining both losses consistently achieves the best performance on both explicit and implicit prompts, demonstrating the effectiveness of our loss design.

\begin{table}[t]
\centering
\caption{Ablation study on the loss function for facial expression prediction.}
\begin{tabular}{l|ccc|ccc}
\toprule
\multirow{2}{*}{Method} 
& \multicolumn{3}{c|}{Explicit} 
& \multicolumn{3}{c}{Impicit} \\
& $L1_{Head}\downarrow$ & $L1_{Face}\downarrow$ & $L1_{Lip}\downarrow$
& $L1_{Head}\downarrow$ & $L1_{Face}\downarrow$ & $L1_{Lip}\downarrow$\\
\midrule
FLAME loss
& 0.62 & 1.02 & 2.42
& 0.75 & 1.14 & 2.85\\
Geo. loss
 & 0.61 & 0.99 &2.32
& 0.67 & 1.13 &2.80 \\
FLAME loss + Geo. loss(Ours)
& \textbf{0.60} & \textbf{0.98} &\textbf{2.29}
& \textbf{0.66} & \textbf{1.09}&\textbf{2.69} \\
\bottomrule
\end{tabular}
\label{tab:ablation_loss}
\end{table}

\section{Runtime of Different Approaches}
Table~\ref{table:running_time} summarizes the runtime of the compared methods, measured on a single L40s GPU. 
Although the models produce outputs in different formats, \OURS exhibits a clear efficiency advantage, generating facial expressions from text substantially faster than competing approaches.

\input{tables/running_time}

\section{Qualitative results of ablation study}
\begin{figure}[h]
  \centering
   \includegraphics[width=\linewidth]{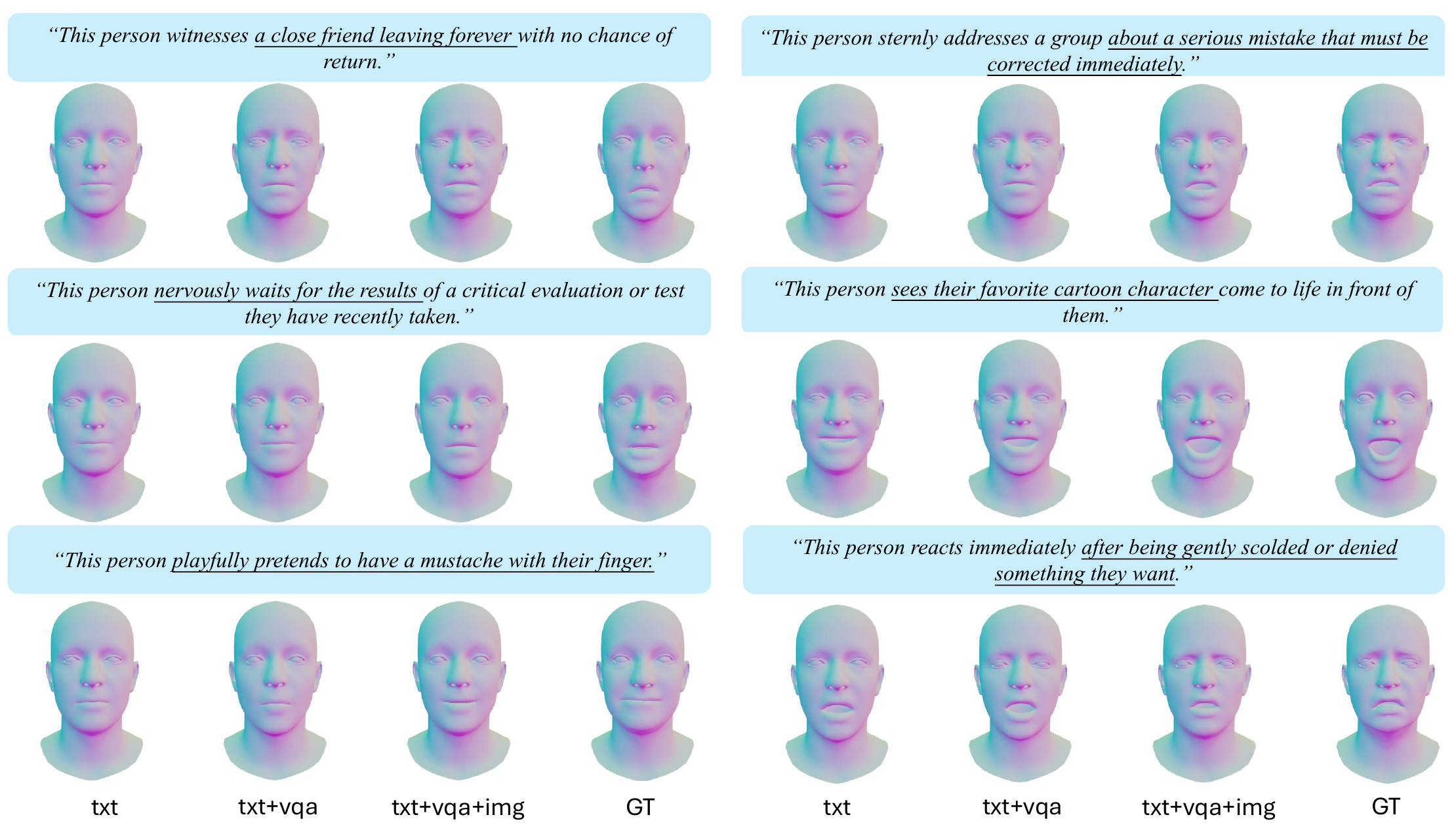}
   \caption{Ablation on the training data. Visual comparison of \OURS trained with different training data on text-to-3D expression synthesis. The inclusion of facial images and VQA data improves expression synthesis, demonstrating how multi-modal inputs enhance the performance for both explicit and implicit-based expressions.}
   \label{fig:ablation_data1}
\end{figure}

Figure~\ref{fig:ablation_data1} illustrates the results of models trained with different data settings. Multimodal training data enable the model to better capture textual semantics and generate more accurate expressions compared to unimodal training.

\begin{figure}[ht]
  \centering
   \includegraphics[width=0.57\linewidth]{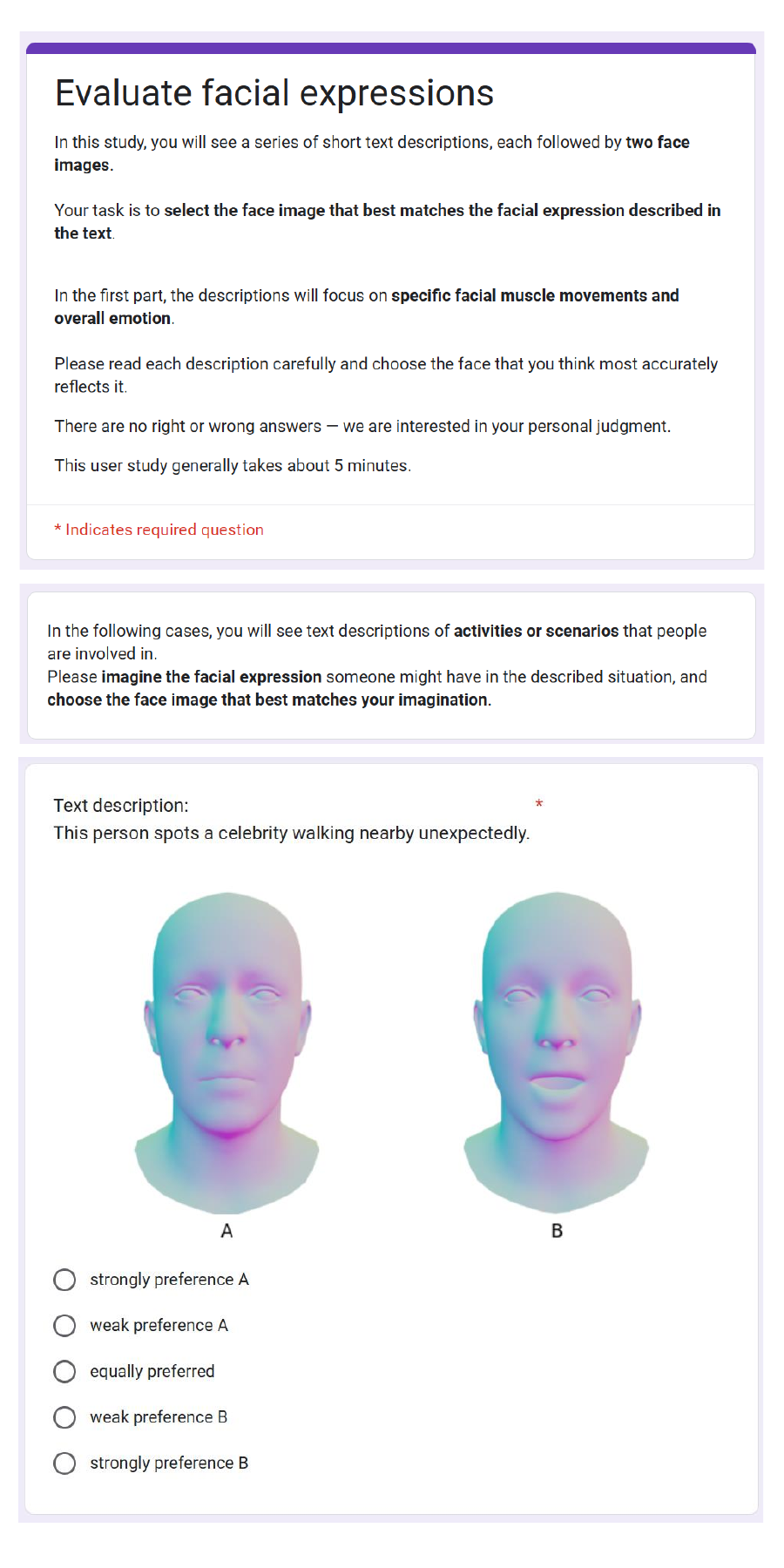}

   \caption{Screenshot of our perceptual study}
   \label{fig:supp:user_study}
\end{figure}

\section{Setting of perceptual study}
We provide the screenshot of our perceptual study in Figure~\ref{fig:supp:user_study}. For a fair comparison, we fit the FLAME model~\cite{FLAME:SiggraphAsia2017} to the prediction of Portrat3D and compare our results with the fitted FLAME model.

\section{Failure Cases}

\begin{figure}[h]
  \centering
   \includegraphics[width=\linewidth]{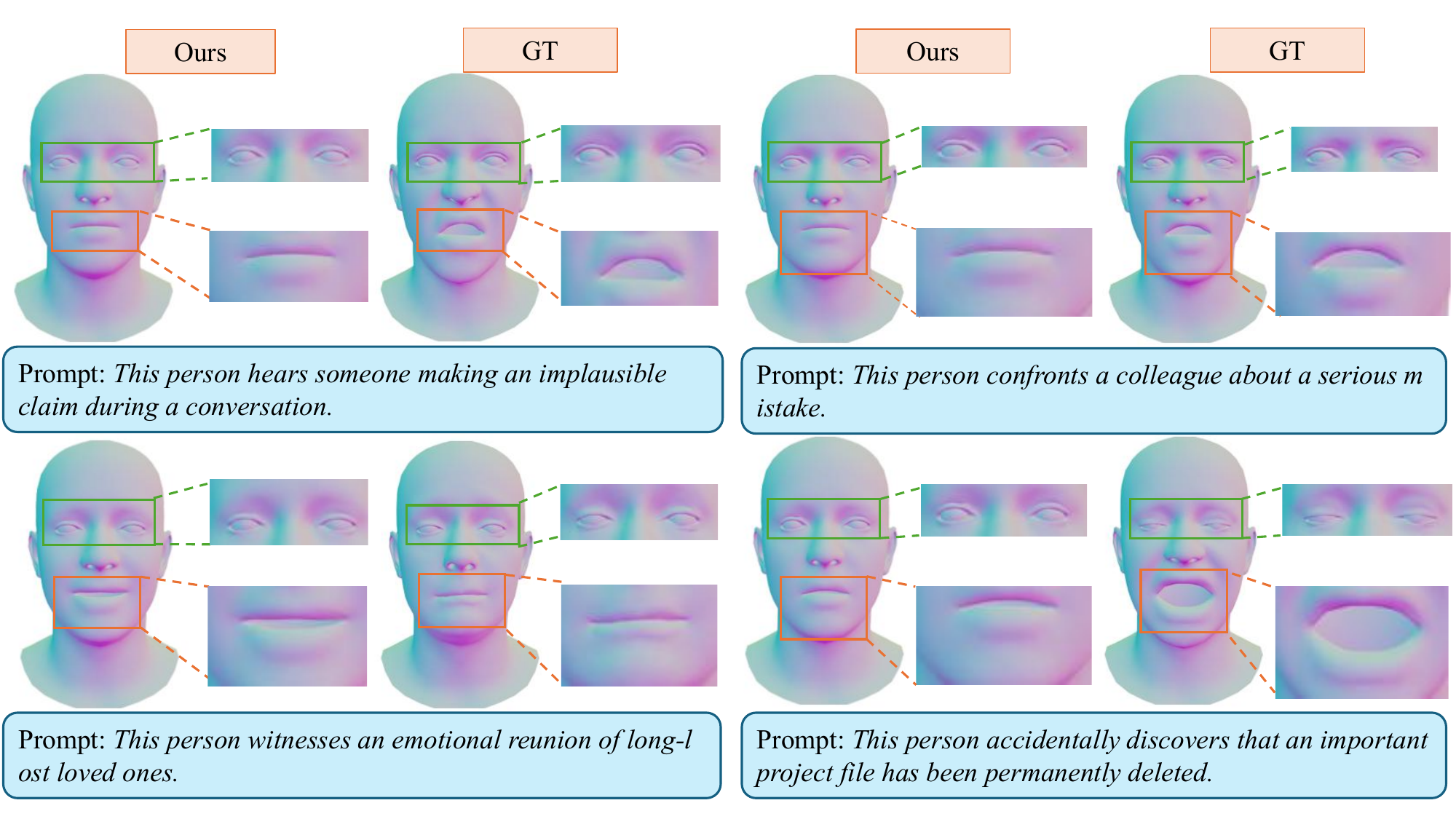}
   \caption{Ablation on the training data. Visual comparison of \OURS trained with different training data on text-to-3D expression synthesis. The inclusion of facial images and VQA data improves expression synthesis, demonstrating how multi-modal inputs enhance the performance for both explicit and implicit-based expressions.}
   \label{fig:fail_cases}
\end{figure}

We provide several failure cases of \OURS in Fig.~\ref{fig:fail_cases}. 
Most failures occur when the prompt is emotionally ambiguous or underspecifies the intended reaction, allowing multiple plausible expressions such as surprise, concern, anger, or sadness. 
In such cases, \OURS may generate a semantically reasonable but more subdued expression, while the ground truth exhibits a stronger reaction. 
We also observe that \OURS may miss fine-grained facial details in complex scenarios, especially around the eyes and mouth.

\section{More qualitative results}

\begin{figure}[h]
  \centering
   \includegraphics[width=\linewidth]{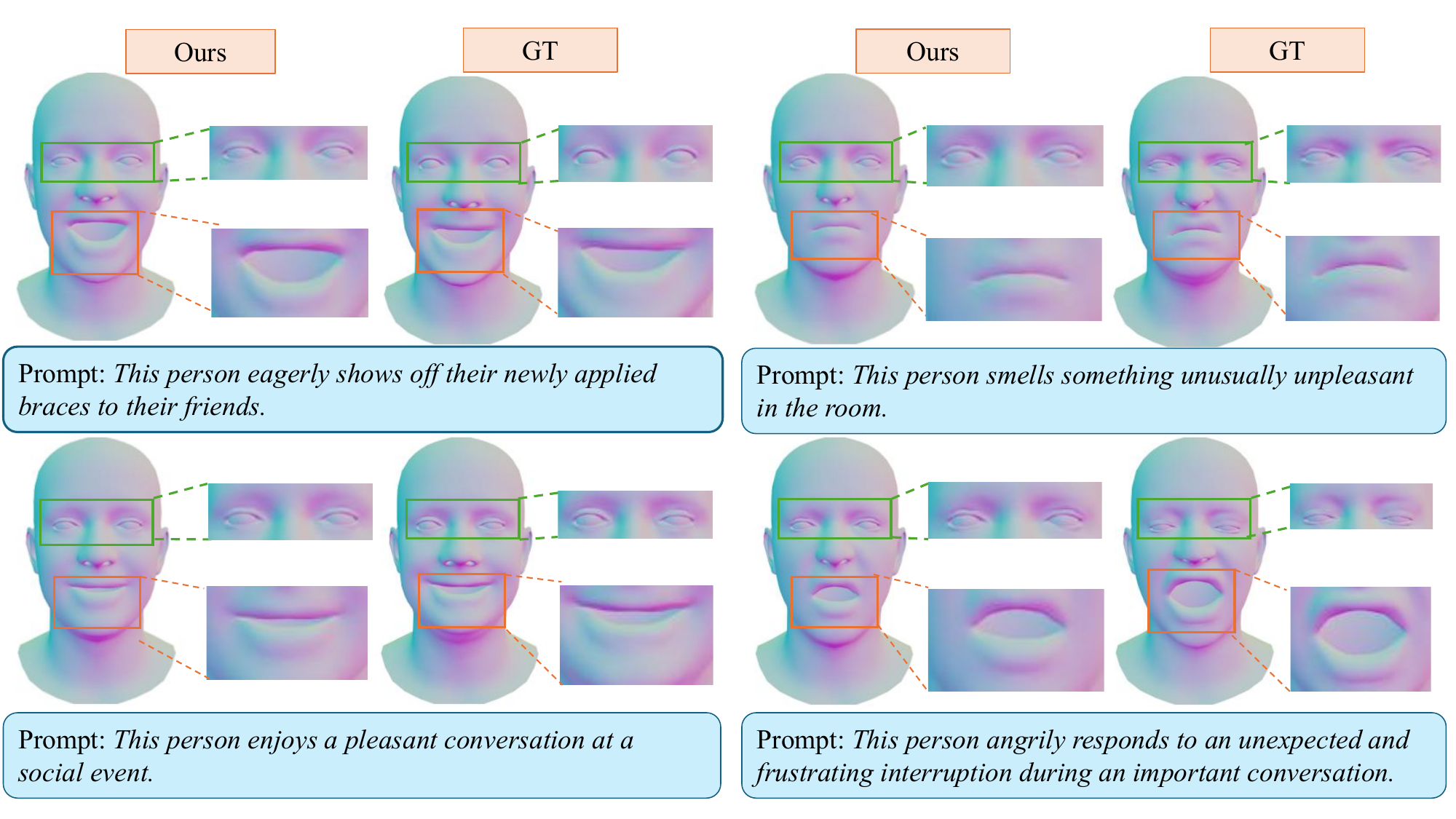}
   \caption{Additional visual comparisons with zoomed-in details. The examples cover diverse prompt-conditioned expressions, including positive and negative affective states. }
   \label{fig:zoomin}
\end{figure}

We provide additional qualitative results with zoomed-in facial details in Fig.~\ref{fig:zoomin}.
The examples show that \OURS can generate diverse expressions conditioned on different textual descriptions. 
Even within the same broad affective category, the predicted expressions vary according to the prompt semantics. 
The zoomed-in regions further show that \OURS captures fine-grained variations around the eyes and mouth, producing results that are generally consistent with the ground-truth expressions. 
These results suggest that our method supports fine-grained expression control beyond coarse emotion categories.

\begin{figure}[ht]
  \centering
   \includegraphics[width=\linewidth]{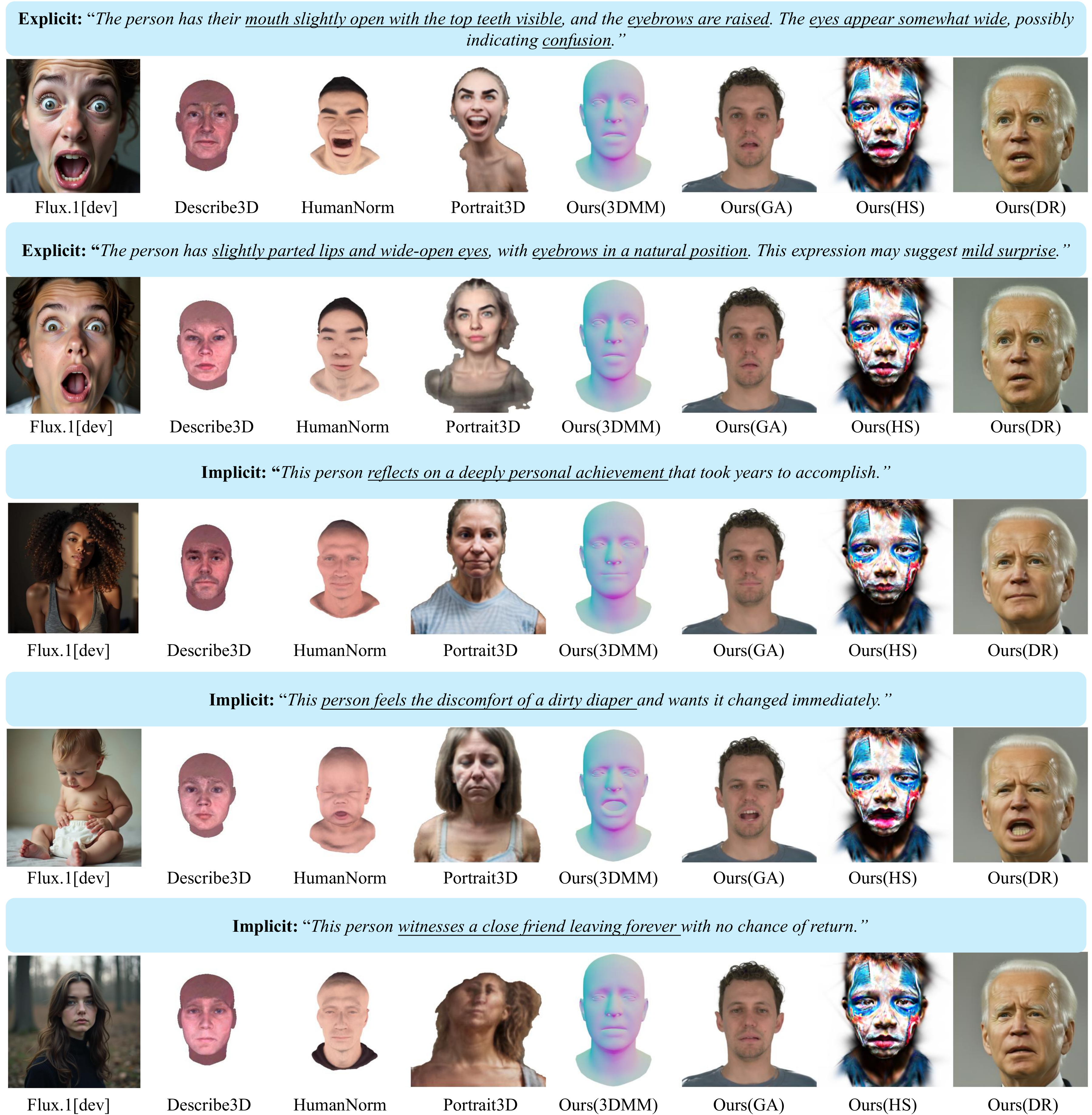}

   \caption{
   Qualitative comparison of outputs from \OURS and other relevant methods. For \OURS, we additionally show results after exporting the generated expressions to GaussianAvatar (GA), HeadStudio (HS), and DiffusionRig (DR). 
   Compared with existing methods, \OURS produces more semantically faithful and expressive results across diverse inputs. It accurately captures subtle composite expressions specified by complex explicit text prompts and also generates expressive results from implicit descriptions.
   }
   \label{fig:vis_res_sys}
\end{figure}

We include additional qualitative comparisons with other methods in Figure~\ref{fig:vis_res_sys}. 
For \textit{explicit} prompts, \OURS better matches fine-grained compositional cues in the text, jointly controlling mouth opening, eye openness, and eyebrow configuration. In contrast, competing methods often overemphasize a single attribute or miss secondary cues, producing expressions that are exaggerated and ambiguous. This is particularly evident in the first two rows, where \OURS more accurately captures nuanced states such as confusion and mild surprise.
For \textit{implicit} prompts, \OURS also infers more plausible affective states from scenario-level descriptions. It produces restrained satisfaction, discomfort, and sadness in the corresponding examples, while several baselines remain near-neutral or inconsistent with the prompt. The exported results on GaussianAvatar, HeadStudio, and DiffusionRig further preserve these semantics, indicating that the expression control generated by \OURS is robust and transferable across rendering backbones.

\begin{figure}[ht]
  \centering
   \includegraphics[width=\textwidth]{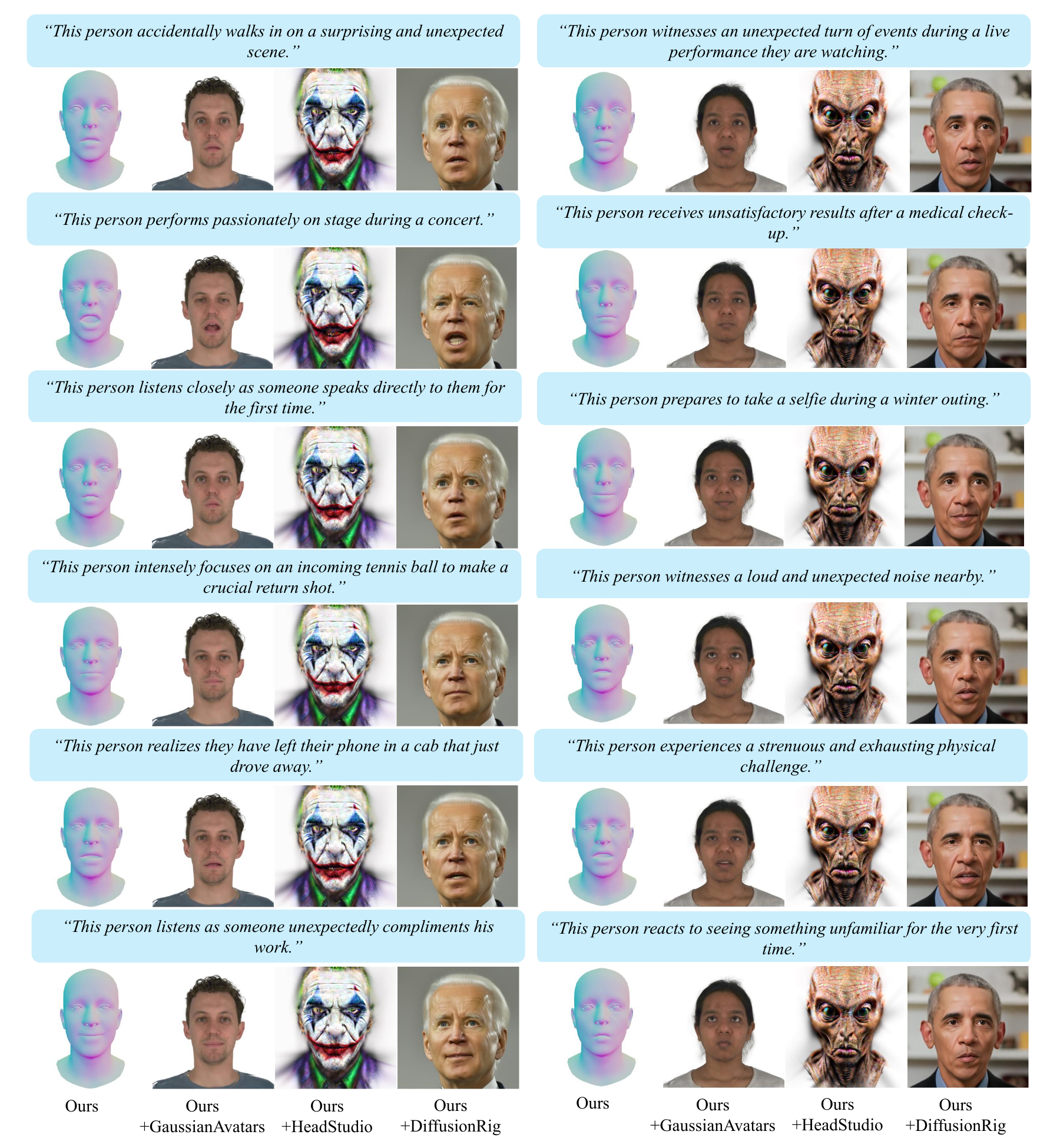}

   \caption{Additional qualitative results generated by EmoteGPT from implicit textual descriptions. EmoteGPT effectively complements existing approaches by synthesizing vivid and realistic facial expressions across a range of contexts.}
   \label{fig:supp:more_applications}
\end{figure}

In Figure~\ref{fig:supp:more_applications}, we also present additional results generated by EmoteGPT in response to a variety of implicit textual inputs. We also include examples where EmoteGPT is combined with other FLAME-based methods. These results present EmoteGPT effectively complements existing approaches, enabling the creation of more expressive 3D avatars and facilitating personalized image generation.

%% file: tables/running_time.tex
\begin{table}[t]
\centering
\caption{Running time of different methods for processing a single prompt. Although prior methods target full 3D head or image synthesis, while \OURS predicts only 3DMM expression parameters, \OURS remains highly efficient and can be readily integrated into 3DMM-based 3D head synthesis pipelines.}
\begin{tabular}{l|c}
\toprule
Method       & Running Time \\ 
\midrule 
SDXL & 4s \\
Flux.1[dev]     &  120s      \\
Describe3D     & 40s      \\
HumanNorm     &  120min      \\
Portrait3D     &   33min    \\
\OURS &  0.5s       \\
\bottomrule
\end{tabular}

\label{table:running_time}
\end{table}

%% file: main.bib
@String(CVPR  = {IEEE Conf. Comput. Vis. Pattern Recog.})

@String(ICCV  = {Int. Conf. Comput. Vis.})

@String(ECCV  = {Eur. Conf. Comput. Vis.})

@String(AAAI  = {AAAI})

@String(TOG   = {ACM Trans. Graph.})

@String(CVPR  = {CVPR})

@String(ICCV  = {ICCV})

@String(ECCV  = {ECCV})

@String(TOG   = {ACM TOG})

@article{egger20203d,
  title={3d morphable face models—past, present, and future},
  author={Egger, Bernhard and Smith, William AP and Tewari, Ayush and Wuhrer, Stefanie and Zollhoefer, Michael and Beeler, Thabo and Bernard, Florian and Bolkart, Timo and Kortylewski, Adam and Romdhani, Sami and others},
  journal={ACM Transactions on Graphics (ToG)},
  volume={39},
  number={5},
  pages={1--38},
  year={2020},
  publisher={ACM New York, NY, USA}
}

@misc{liu2023improvedllava,
          author={Liu, Haotian and Li, Chunyuan and Li, Yuheng and Lee, Yong Jae},
          title={Improved Baselines with Visual Instruction Tuning}, 
          publisher={arXiv:2310.03744},
          year={2023},
  }

@inproceedings{blanz1999morphable,
  title={A Morphable Model for the Synthesis of 3D Faces},
  author={Blanz, V and Vetter, T},
  booktitle={26th Annual Conference on Computer Graphics and Interactive Techniques (SIGGRAPH 1999)},
  pages={187--194},
  year={1999},
  organization={ACM Press}
}

@inproceedings{aneja2023clipface, series={SIGGRAPH ’23},
   title={ClipFace: Text-guided Editing of Textured 3D Morphable Models},
   url={http://dx.doi.org/10.1145/3588432.3591566},
   DOI={10.1145/3588432.3591566},
   booktitle={Special Interest Group on Computer Graphics and Interactive Techniques Conference Conference Proceedings},
   publisher={ACM},
   author={Aneja, Shivangi and Thies, Justus and Dai, Angela and Niessner, Matthias},
   year={2023},
   month=jul, pages={1–11},
   collection={SIGGRAPH ’23} }

@misc{magicmirrorfasthighqualityavatar2024,
      title={MagicMirror: Fast and High-Quality Avatar Generation with a Constrained Search Space}, 
      author={Armand Comas-Massagué and Di Qiu and Menglei Chai and Marcel Bühler and Amit Raj and Ruiqi Gao and Qiangeng Xu and Mark Matthews and Paulo Gotardo and Octavia Camps and Sergio Orts-Escolano and Thabo Beeler},
      year={2024},
      eprint={2404.01296},
      archivePrefix={arXiv},
      primaryClass={cs.CV},
      url={https://arxiv.org/abs/2404.01296}, 
}

@article{10.1145/3618368,
author = {Mendiratta, Mohit and Pan, Xingang and Elgharib, Mohamed and Teotia, Kartik and R, Mallikarjun B and Tewari, Ayush and Golyanik, Vladislav and Kortylewski, Adam and Theobalt, Christian},
title = {AvatarStudio: Text-Driven Editing of 3D Dynamic Human Head Avatars},
year = {2023},
issue_date = {December 2023},
publisher = {Association for Computing Machinery},
address = {New York, NY, USA},
volume = {42},
number = {6},
issn = {0730-0301},
url = {https://doi.org/10.1145/3618368},
doi = {10.1145/3618368},
journal = {ACM Trans. Graph.},
month = {dec},
articleno = {226},
numpages = {18}
}

@article{wei2023dediffusion,
      author    = {Wei, Chen and Liu, Chenxi and Qiao, Siyuan and Zhang, Zhishuai and Yuille, Alan and Yu, Jiahui},
      title     = {De-Diffusion Makes Text a Strong Cross-Modal Interface},
      journal   = {arXiv preprint arXiv:2311.00618},
      year      = {2023},
    }

@InProceedings{lai2023lisa,
  title={LISA: Reasoning Segmentation via Large Language Model},
  author={Lai, Xin and Tian, Zhuotao and Chen, Yukang and Li, Yanwei and Yuan, Yuhui and Liu, Shu and Jia, Jiaya},
  booktitle={CVPR},
  year={2024}
}

@InProceedings{feng2024chatpose,
    author = {Feng, Yao and Lin, Jing and Dwivedi, Sai Kumar and Sun, Yu and Patel, Priyanka and Black, Michael J.},
    title = {ChatPose: Chatting about 3D Human Pose},
    booktitle={CVPR},
    year={2024}
}

@inproceedings{
hu2022lora,
title={Lo{RA}: Low-Rank Adaptation of Large Language Models},
author={Edward J Hu and Yelong Shen and Phillip Wallis and Zeyuan Allen-Zhu and Yuanzhi Li and Shean Wang and Lu Wang and Weizhu Chen},
booktitle={International Conference on Learning Representations},
year={2022},
url={https://openreview.net/forum?id=nZeVKeeFYf9}
}

@inproceedings{bulat2017far,
  title={How far are we from solving the 2D \& 3D Face Alignment problem? (and a dataset of 230,000 3D facial landmarks)},
  author={Bulat, Adrian and Tzimiropoulos, Georgios},
  booktitle={International Conference on Computer Vision},
  year={2017}
}

@inproceedings{raj2020zeroopt,
author = {Rajbhandari, Samyam and Rasley, Jeff and Ruwase, Olatunji and He, Yuxiong},
title = {ZeRO: memory optimizations toward training trillion parameter models},
year = {2020},
isbn = {9781728199986},
publisher = {IEEE Press},
booktitle = {Proceedings of the International Conference for High Performance Computing, Networking, Storage and Analysis},
articleno = {20},
numpages = {16},
location = {Atlanta, Georgia},
series = {SC '20}
}

@inproceedings{ras2020deepspeed,
author = {Rasley, Jeff and Rajbhandari, Samyam and Ruwase, Olatunji and He, Yuxiong},
title = {DeepSpeed: System Optimizations Enable Training Deep Learning Models with Over 100 Billion Parameters},
year = {2020},
isbn = {9781450379984},
publisher = {Association for Computing Machinery},
address = {New York, NY, USA},
url = {https://doi.org/10.1145/3394486.3406703},
doi = {10.1145/3394486.3406703},
booktitle = {Proceedings of the 26th ACM SIGKDD International Conference on Knowledge Discovery \& Data Mining},
pages = {3505–3506},
numpages = {2},
location = {Virtual Event, CA, USA},
series = {KDD '20}
}

@inproceedings{
loshchilov2018adamw,
title={Decoupled Weight Decay Regularization},
author={Ilya Loshchilov and Frank Hutter},
booktitle={International Conference on Learning Representations},
year={2019},
url={https://openreview.net/forum?id=Bkg6RiCqY7},
}

@article{zhu2023minigpt,
  title={MiniGPT-4: Enhancing Vision-Language Understanding with Advanced Large Language Models},
  author={Zhu, Deyao and Chen, Jun and Shen, Xiaoqian and Li, Xiang and Elhoseiny, Mohamed},
  journal={arXiv preprint arXiv:2304.10592},
  year={2023}
}

@inproceedings{danvevcek2022emoca,
  title={Emoca: Emotion driven monocular face capture and animation},
  author={Dan{\v{e}}{\v{c}}ek, Radek and Black, Michael J and Bolkart, Timo},
  booktitle={Proceedings of the IEEE/CVF Conference on Computer Vision and Pattern Recognition},
  pages={20311--20322},
  year={2022}
}

@article{hendrycksG2016gelu,
  author       = {Dan Hendrycks and
                  Kevin Gimpel},
  title        = {Bridging Nonlinearities and Stochastic Regularizers with Gaussian
                  Error Linear Units},
  journal      = {CoRR},
  volume       = {abs/1606.08415},
  year         = {2016},
  url          = {http://arxiv.org/abs/1606.08415},
  eprinttype    = {arXiv},
  eprint       = {1606.08415},
  bibsource    = {dblp computer science bibliography, https://dblp.org}
}

@misc{gpt4,
      title={GPT-4 Technical Report}, 
      author={OpenAI and Josh Achiam and Steven Adler and Sandhini Agarwal and Lama Ahmad and Ilge Akkaya and Florencia Leoni Aleman and Diogo Almeida and Janko Altenschmidt and Sam Altman and Shyamal Anadkat and Red Avila and Igor Babuschkin and Suchir Balaji and Valerie Balcom and Paul Baltescu and Haiming Bao and Mohammad Bavarian and Jeff Belgum and Irwan Bello and Jake Berdine and Gabriel Bernadett-Shapiro and Christopher Berner and Lenny Bogdonoff and Oleg Boiko and Madelaine Boyd and Anna-Luisa Brakman and Greg Brockman and Tim Brooks and Miles Brundage and Kevin Button and Trevor Cai and Rosie Campbell and Andrew Cann and Brittany Carey and Chelsea Carlson and Rory Carmichael and Brooke Chan and Che Chang and Fotis Chantzis and Derek Chen and Sully Chen and Ruby Chen and Jason Chen and Mark Chen and Ben Chess and Chester Cho and Casey Chu and Hyung Won Chung and Dave Cummings and Jeremiah Currier and Yunxing Dai and Cory Decareaux and Thomas Degry and Noah Deutsch and Damien Deville and Arka Dhar and David Dohan and Steve Dowling and Sheila Dunning and Adrien Ecoffet and Atty Eleti and Tyna Eloundou and David Farhi and Liam Fedus and Niko Felix and Simón Posada Fishman and Juston Forte and Isabella Fulford and Leo Gao and Elie Georges and Christian Gibson and Vik Goel and Tarun Gogineni and Gabriel Goh and Rapha Gontijo-Lopes and Jonathan Gordon and Morgan Grafstein and Scott Gray and Ryan Greene and Joshua Gross and Shixiang Shane Gu and Yufei Guo and Chris Hallacy and Jesse Han and Jeff Harris and Yuchen He and Mike Heaton and Johannes Heidecke and Chris Hesse and Alan Hickey and Wade Hickey and Peter Hoeschele and Brandon Houghton and Kenny Hsu and Shengli Hu and Xin Hu and Joost Huizinga and Shantanu Jain and Shawn Jain and Joanne Jang and Angela Jiang and Roger Jiang and Haozhun Jin and Denny Jin and Shino Jomoto and Billie Jonn and Heewoo Jun and Tomer Kaftan and Łukasz Kaiser and Ali Kamali and Ingmar Kanitscheider and Nitish Shirish Keskar and Tabarak Khan and Logan Kilpatrick and Jong Wook Kim and Christina Kim and Yongjik Kim and Jan Hendrik Kirchner and Jamie Kiros and Matt Knight and Daniel Kokotajlo and Łukasz Kondraciuk and Andrew Kondrich and Aris Konstantinidis and Kyle Kosic and Gretchen Krueger and Vishal Kuo and Michael Lampe and Ikai Lan and Teddy Lee and Jan Leike and Jade Leung and Daniel Levy and Chak Ming Li and Rachel Lim and Molly Lin and Stephanie Lin and Mateusz Litwin and Theresa Lopez and Ryan Lowe and Patricia Lue and Anna Makanju and Kim Malfacini and Sam Manning and Todor Markov and Yaniv Markovski and Bianca Martin and Katie Mayer and Andrew Mayne and Bob McGrew and Scott Mayer McKinney and Christine McLeavey and Paul McMillan and Jake McNeil and David Medina and Aalok Mehta and Jacob Menick and Luke Metz and Andrey Mishchenko and Pamela Mishkin and Vinnie Monaco and Evan Morikawa and Daniel Mossing and Tong Mu and Mira Murati and Oleg Murk and David Mély and Ashvin Nair and Reiichiro Nakano and Rajeev Nayak and Arvind Neelakantan and Richard Ngo and Hyeonwoo Noh and Long Ouyang and Cullen O'Keefe and Jakub Pachocki and Alex Paino and Joe Palermo and Ashley Pantuliano and Giambattista Parascandolo and Joel Parish and Emy Parparita and Alex Passos and Mikhail Pavlov and Andrew Peng and Adam Perelman and Filipe de Avila Belbute Peres and Michael Petrov and Henrique Ponde de Oliveira Pinto and Michael and Pokorny and Michelle Pokrass and Vitchyr H. Pong and Tolly Powell and Alethea Power and Boris Power and Elizabeth Proehl and Raul Puri and Alec Radford and Jack Rae and Aditya Ramesh and Cameron Raymond and Francis Real and Kendra Rimbach and Carl Ross and Bob Rotsted and Henri Roussez and Nick Ryder and Mario Saltarelli and Ted Sanders and Shibani Santurkar and Girish Sastry and Heather Schmidt and David Schnurr and John Schulman and Daniel Selsam and Kyla Sheppard and Toki Sherbakov and Jessica Shieh and Sarah Shoker and Pranav Shyam and Szymon Sidor and Eric Sigler and Maddie Simens and Jordan Sitkin and Katarina Slama and Ian Sohl and Benjamin Sokolowsky and Yang Song and Natalie Staudacher and Felipe Petroski Such and Natalie Summers and Ilya Sutskever and Jie Tang and Nikolas Tezak and Madeleine B. Thompson and Phil Tillet and Amin Tootoonchian and Elizabeth Tseng and Preston Tuggle and Nick Turley and Jerry Tworek and Juan Felipe Cerón Uribe and Andrea Vallone and Arun Vijayvergiya and Chelsea Voss and Carroll Wainwright and Justin Jay Wang and Alvin Wang and Ben Wang and Jonathan Ward and Jason Wei and CJ Weinmann and Akila Welihinda and Peter Welinder and Jiayi Weng and Lilian Weng and Matt Wiethoff and Dave Willner and Clemens Winter and Samuel Wolrich and Hannah Wong and Lauren Workman and Sherwin Wu and Jeff Wu and Michael Wu and Kai Xiao and Tao Xu and Sarah Yoo and Kevin Yu and Qiming Yuan and Wojciech Zaremba and Rowan Zellers and Chong Zhang and Marvin Zhang and Shengjia Zhao and Tianhao Zheng and Juntang Zhuang and William Zhuk and Barret Zoph},
      year={2024},
      eprint={2303.08774},
      archivePrefix={arXiv},
      primaryClass={cs.CL}
}

@article{liu2024visual,
  title={Visual instruction tuning},
  author={Liu, Haotian and Li, Chunyuan and Wu, Qingyang and Lee, Yong Jae},
  journal={Advances in neural information processing systems},
  volume={36},
  year={2024}
}

@article{su2023pandagpt,
  title={Pandagpt: One model to instruction-follow them all},
  author={Su, Yixuan and Lan, Tian and Li, Huayang and Xu, Jialu and Wang, Yan and Cai, Deng},
  journal={arXiv preprint arXiv:2305.16355},
  year={2023}
}

@inproceedings{girdhar2023imagebind,
  title={Imagebind: One embedding space to bind them all},
  author={Girdhar, Rohit and El-Nouby, Alaaeldin and Liu, Zhuang and Singh, Mannat and Alwala, Kalyan Vasudev and Joulin, Armand and Misra, Ishan},
  booktitle={Proceedings of the IEEE/CVF Conference on Computer Vision and Pattern Recognition},
  pages={15180--15190},
  year={2023}
}

@article{wu2023next,
  title={Next-gpt: Any-to-any multimodal llm},
  author={Wu, Shengqiong and Fei, Hao and Qu, Leigang and Ji, Wei and Chua, Tat-Seng},
  journal={arXiv preprint arXiv:2309.05519},
  year={2023}
}

@InProceedings{ding2023diffusionrig,
      author    = {Zheng, Ding and Cecilia, Zhang and Zhihao, Xia and Lars, Jebe and Zhuowen, Tu and Xiuming, Zhang},
      title     = {DiffusionRig: Learning Personalized Priors for Facial Appearance Editing},
      booktitle = {Proceedings of the IEEE/CVF Conference on Computer Vision and Pattern Recognition},
      year      = {2023},
}

@inproceedings{liu2015celeba,
  title = {Deep Learning Face Attributes in the Wild},
  author = {Liu, Ziwei and Luo, Ping and Wang, Xiaogang and Tang, Xiaoou},
  booktitle = {Proceedings of International Conference on Computer Vision (ICCV)},
  month = {December},
  year = {2015} 
}

@inproceedings{celebatext2021,
author = {Sun, Jianxin and Li, Qi and Wang, Weining and Zhao, Jian and Sun, Zhenan},
title = {Multi-caption Text-to-Face Synthesis: Dataset and Algorithm},
year = {2021},
isbn = {9781450386517},
publisher = {Association for Computing Machinery},
address = {New York, NY, USA},
url = {https://doi.org/10.1145/3474085.3475391},
doi = {10.1145/3474085.3475391},
booktitle = {Proceedings of the 29th ACM International Conference on Multimedia},
pages = {2290–2298},
numpages = {9},
location = {Virtual Event, China},
series = {MM '21}
}

@INPROCEEDINGS {describe3d2023,
author = {M. Wu and H. Zhu and L. Huang and Y. Zhuang and Y. Lu and X. Cao},
booktitle = {2023 IEEE/CVF Conference on Computer Vision and Pattern Recognition (CVPR)},
title = {High-fidelity 3D Face Generation from Natural Language Descriptions},
year = {2023},
volume = {},
issn = {},
pages = {4521-4530},
doi = {10.1109/CVPR52729.2023.00439},
url = {https://doi.ieeecomputersociety.org/10.1109/CVPR52729.2023.00439},
publisher = {IEEE Computer Society},
address = {Los Alamitos, CA, USA},
month = {jun}
}

@article{FLAME:SiggraphAsia2017, 
  title = {Learning a model of facial shape and expression from {4D} scans}, 
  author = {Li, Tianye and Bolkart, Timo and Black, Michael. J. and Li, Hao and Romero, Javier}, 
  journal = {ACM Transactions on Graphics, (Proc. SIGGRAPH Asia)}, 
  volume = {36}, 
  number = {6}, 
  year = {2017}, 
  pages = {194:1--194:17},
  url = {https://doi.org/10.1145/3130800.3130813} 
}

@article{affectnet2019,
author = {Mollahosseini, Ali and Hasani, Behzad and Mahoor, Mohammad H.},
title = {AffectNet: A Database for Facial Expression, Valence, and Arousal Computing in the Wild},
year = {2019},
issue_date = {January 2019},
publisher = {IEEE Computer Society Press},
address = {Washington, DC, USA},
volume = {10},
number = {1},
issn = {1949-3045},
url = {https://doi.org/10.1109/TAFFC.2017.2740923},
doi = {10.1109/TAFFC.2017.2740923},
journal = {IEEE Trans. Affect. Comput.},
month = jan,
pages = {18–31},
numpages = {14}
}

@inproceedings{qian2024gaussianavatars,
  title={Gaussianavatars: Photorealistic head avatars with rigged 3d gaussians},
  author={Qian, Shenhan and Kirschstein, Tobias and Schoneveld, Liam and Davoli, Davide and Giebenhain, Simon and Nie{\ss}ner, Matthias},
  booktitle={Proceedings of the IEEE/CVF Conference on Computer Vision and Pattern Recognition},
  pages={20299--20309},
  year={2024}
}

@inproceedings{sig2023cliphead,
author = {Manu, Pranav and Srivastava, Astitva and Sharma, Avinash},
title = {CLIP-Head: Text-Guided Generation of Textured Neural Parametric 3D Head Models},
year = {2023},
isbn = {9798400703140},
publisher = {Association for Computing Machinery},
address = {New York, NY, USA},
url = {https://doi.org/10.1145/3610543.3626169},
doi = {10.1145/3610543.3626169},
booktitle = {SIGGRAPH Asia 2023 Technical Communications},
articleno = {29},
numpages = {4},
location = {Sydney, NSW, Australia},
series = {SA '23}
}

@article{toisoul2021emonet,
  author  = {Antoine Toisoul and Jean Kossaifi and Adrian Bulat and Georgios Tzimiropoulos and Maja Pantic},
  title   = {Estimation of continuous valence and arousal levels from faces in naturalistic conditions},
  journal = {Nature Machine Intelligence},
  year    = {2021},
  url     = {https://www.nature.com/articles/s42256-020-00280-0}
}

@inproceedings{zhou2024headstudio,
  title = {HeadStudio: Text to Animatable Head Avatars with 3D Gaussian Splatting},
  author = {Zhenglin Zhou and Fan Ma and Hehe Fan and Zongxin Yang and Yi Yang},
  booktile = {ECCV},
  year = {2024},
}

@inproceedings{neurips2023vicuna,
author = {Zheng, Lianmin and Chiang, Wei-Lin and Sheng, Ying and Zhuang, Siyuan and Wu, Zhanghao and Zhuang, Yonghao and Lin, Zi and Li, Zhuohan and Li, Dacheng and Xing, Eric P. and Zhang, Hao and Gonzalez, Joseph E. and Stoica, Ion},
title = {Judging LLM-as-a-judge with MT-bench and Chatbot Arena},
year = {2023},
publisher = {Curran Associates Inc.},
address = {Red Hook, NY, USA},
booktitle = {Proceedings of the 37th International Conference on Neural Information Processing Systems},
articleno = {2020},
numpages = {29},
location = {New Orleans, LA, USA},
series = {NIPS '23}
}

@inproceedings{
poole2023dreamfusion,
title={DreamFusion: Text-to-3D using 2D Diffusion},
author={Ben Poole and Ajay Jain and Jonathan T. Barron and Ben Mildenhall},
booktitle={The Eleventh International Conference on Learning Representations },
year={2023},
url={https://openreview.net/forum?id=FjNys5c7VyY}
}

@misc{huang2023humannorm,
title={Humannorm: Learning normal diffusion model for high-quality and realistic 3d human generation},
author={Huang, Xin and Shao, Ruizhi and Zhang, Qi and Zhang, Hongwen and Feng, Ying and Liu, Yebin and Wang, Qing},
booktitle={Proceedings of the IEEE Conference on Computer Vision and Pattern Recognition},
year={2024}
}

@article{Portrait3D_sig24,
author = {Wu, Yiqian and Xu, Hao and Tang, Xiangjun and Chen, Xien and Tang, Siyun and Zhang, Zhebin and Li, Chen and Jin, Xiaogang},
title = {Portrait3D: Text-Guided High-Quality 3D Portrait Generation Using Pyramid Representation and GANs Prior},
year = {2024},
publisher = {Association for Computing Machinery},
volume = {43},
number = {4},
url = {https://doi.org/10.1145/3658162},
doi = {10.1145/3658162},
journal = {ACM Trans. Graph.},
month = {Jul},
articleno = {45}
}

@misc{retsinas20243dfacialexpressionsanalysisbyneuralsynthesis,
      title={3D Facial Expressions through Analysis-by-Neural-Synthesis}, 
      author={George Retsinas and Panagiotis P. Filntisis and Radek Danecek and Victoria F. Abrevaya and Anastasios Roussos and Timo Bolkart and Petros Maragos},
      year={2024},
      eprint={2404.04104},
      archivePrefix={arXiv},
      primaryClass={cs.CV},
      url={https://arxiv.org/abs/2404.04104}, 
}

@inproceedings{zielonka2022metricalreconstructionhumanfaces,
  title        = {Towards Metrical Reconstruction of Human Faces},
  author       = {Wojciech Zielonka and Timo Bolkart and Justus Thies},
  booktitle    = {ECCV},
  year         = {2022}
}

@ARTICLE{cvpr2019stylegan,
author={Karras, Tero and Laine, Samuli and Aila, Timo},
journal={ IEEE Transactions on Pattern Analysis \& Machine Intelligence },
title={{ A Style-Based Generator Architecture for Generative Adversarial Networks }},
year={2021},
volume={43},
number={12},
ISSN={1939-3539},
pages={4217-4228},
doi={10.1109/TPAMI.2020.2970919},
url = {https://doi.ieeecomputersociety.org/10.1109/TPAMI.2020.2970919},
publisher={IEEE Computer Society},
address={Los Alamitos, CA, USA},
month=dec}

@inproceedings{
podell2024sdxl,
title={{SDXL}: Improving Latent Diffusion Models for High-Resolution Image Synthesis},
author={Dustin Podell and Zion English and Kyle Lacey and Andreas Blattmann and Tim Dockhorn and Jonas M{\"u}ller and Joe Penna and Robin Rombach},
booktitle={The Twelfth International Conference on Learning Representations},
year={2024},
url={https://openreview.net/forum?id=di52zR8xgf}
}

@InProceedings{CelebA-Dialog,
  title = {Talk-to-Edit: Fine-Grained Facial Editing via Dialog},
  author = {Jiang, Yuming and Huang, Ziqi and Pan, Xingang and Loy, Chen Change and Liu, Ziwei},
  booktitle = {Proceedings of the IEEE/CVF International Conference on Computer Vision},
  year={2021}
}

@inproceedings{xia2021tedigan,
  title={TediGAN: Text-Guided Diverse Face Image Generation and Manipulation},
  author={Xia, Weihao and Yang, Yujiu and Xue, Jing-Hao and Wu, Baoyuan},
  booktitle={IEEE Conference on Computer Vision and Pattern Recognition (CVPR)},
  year={2021}
}

@INPROCEEDINGS{ffhqtext,
  author={Zhou, Yutong and Shimada, Nobutaka},
  booktitle={2021 16th IEEE International Conference on Automatic Face and Gesture Recognition (FG 2021)}, 
  title={Generative Adversarial Network for Text-to-Face Synthesis and Manipulation with Pretrained BERT Model}, 
  year={2021},
  volume={},
  number={},
  pages={01-08},
  doi={10.1109/FG52635.2021.9666791}}

@misc{flux2024,
    author={Black Forest Labs},
    title={FLUX},
    year={2024},
    howpublished={\url{https://github.com/black-forest-labs/flux}},
}

@INPROCEEDINGS {IdExpAmbiguityFlame2021,
author = { Egger, Bernhard and Sutherland, Skylar and Medin, Safa C. and Tenenbaum, Joshua },
booktitle = { 2021 16th IEEE International Conference on Automatic Face and Gesture Recognition (FG 2021) },
title = {{ Identity-Expression Ambiguity in 3D Morphable Face Models }},
year = {2021},
volume = {},
ISSN = {},
pages = {1-7},
doi = {10.1109/FG52635.2021.9667002},
url = {https://doi.ieeecomputersociety.org/10.1109/FG52635.2021.9667002},
publisher = {IEEE Computer Society},
address = {Los Alamitos, CA, USA},
month =Dec}

@inproceedings{Rot6d_2019_CVPR,
title={On the Continuity of Rotation Representations in Neural Networks},
author={Zhou, Yi and Barnes, Connelly and Jingwan, Lu and Jimei, Yang and Hao, Li},
booktitle={The IEEE Conference on Computer Vision and Pattern Recognition (CVPR)},
month={June},
year={2019}
}

@inproceedings{EMOTE,
  title = {Emotional Speech-Driven Animation with Content-Emotion Disentanglement},
  author = {Daněček, Radek and Chhatre, Kiran and Tripathi, Shashank and Wen, Yandong and Black, Michael and Bolkart, Timo},
  publisher = {ACM},
  month = dec,
  year = {2023},
  doi = {10.1145/3610548.3618183},
  url = {https://emote.is.tue.mpg.de/index.html},
  month_numeric = {12}
}

@inproceedings{tan2024style2talker,
  title={Style2Talker: High-Resolution Talking Head Generation with Emotion Style and Art Style},
  author={Tan, Shuai and Ji, Bin and Pan, Ye},
  booktitle={Proceedings of the AAAI Conference on Artificial Intelligence},
  volume={38},
  number={5},
  pages={5079--5087},
  year={2024}
}

@misc{ma2024talkcliptalkingheadgeneration,
      title={TalkCLIP: Talking Head Generation with Text-Guided Expressive Speaking Styles}, 
      author={Yifeng Ma and Suzhen Wang and Yu Ding and Bowen Ma and Tangjie Lv and Changjie Fan and Zhipeng Hu and Zhidong Deng and Xin Yu},
      year={2024},
      eprint={2304.00334},
      archivePrefix={arXiv},
      primaryClass={cs.CV},
      url={https://arxiv.org/abs/2304.00334}, 
}

@inproceedings{expclip2024,
author = {Zhong, Yicheng and Wei, Huawei and Yang, Peiji and Wang, Zhisheng},
title = {ExpCLIP: bridging text and facial expressions via semantic alignment},
year = {2024},
isbn = {978-1-57735-887-9},
publisher = {AAAI Press},
url = {https://doi.org/10.1609/aaai.v38i7.28594},
doi = {10.1609/aaai.v38i7.28594},
booktitle = {Proceedings of the Thirty-Eighth AAAI Conference on Artificial Intelligence and Thirty-Sixth Conference on Innovative Applications of Artificial Intelligence and Fourteenth Symposium on Educational Advances in Artificial Intelligence},
articleno = {846},
numpages = {9},
series = {AAAI'24/IAAI'24/EAAI'24}
}

@misc{prinzler2024joker,
      title={Joker: Conditional 3D Head Synthesis with Extreme Facial Expressions},
      author={Malte Prinzler and Egor Zakharov and Vanessa Sklyarova and Berna Kabadayi and Justus Thies},
      year={2024},
      eprint={2410.16395},
      archivePrefix={arXiv},
      primaryClass={cs.CV},
      url={https://arxiv.org/abs/2410.16395},
}

@inproceedings{taubner2025cap4d,
    author    = {Taubner, Felix and Zhang, Ruihang and Tuli, Mathieu and Lindell, David B.},
    title     = {{CAP4D}: Creating Animatable {4D} Portrait Avatars with Morphable Multi-View Diffusion Models},
    booktitle = {Proceedings of the IEEE/CVF Conference on Computer Vision and Pattern Recognition (CVPR)},
    month     = {June},
    year      = {2025},
    pages     = {5318-5330}
}

@misc{taubner2025mvp4d,
  title={{MVP4D}: Multi-View Portrait Video Diffusion for Animatable {4D} Avatars}, 
  author={Felix Taubner and Ruihang Zhang and Mathieu Tuli and Sherwin Bahmani and David B. Lindell},
  year={2025},
  eprint={2510.12785},
  archivePrefix={arXiv},
  primaryClass={cs.CV},
  url={https://arxiv.org/abs/2510.12785}, 
}

@article{team2025zimage,
  title={Z-Image: An Efficient Image Generation Foundation Model with Single-Stream Diffusion Transformer},
  author={Z-Image Team},
  journal={arXiv preprint arXiv:2511.22699},
  year={2025}
}

@inproceedings{lsh1998,
author = {Indyk, Piotr and Motwani, Rajeev},
title = {Approximate nearest neighbors: towards removing the curse of dimensionality},
year = {1998},
isbn = {0897919629},
publisher = {Association for Computing Machinery},
address = {New York, NY, USA},
url = {https://doi.org/10.1145/276698.276876},
doi = {10.1145/276698.276876},
booktitle = {Proceedings of the Thirtieth Annual ACM Symposium on Theory of Computing},
pages = {604–613},
numpages = {10},
location = {Dallas, Texas, USA},
series = {STOC '98}
}
